\documentclass[preprint,3p, twocolumn]{elsarticle}

\usepackage{amssymb}
\usepackage{multirow}
\usepackage{amsmath}

\usepackage{color}
\DeclareMathOperator{\sign}{sign}
\DeclareMathOperator*{\argmax}{argmax}
\DeclareMathOperator*{\argmin}{argmin}
\usepackage{algorithm}
\usepackage{algorithmic}
\usepackage{xcolor} 
\usepackage{ulem} 

\newcommand{\authone}[1]{\textcolor{black}{#1}}  
\newcommand{\authtwo}[1]{\textcolor{black}{#1}} 
\newcommand{\authfour}[1]{\textcolor{black}{#1}} 
\newcommand{\geluconst}{0.044715}

\begin{document}

\begin{frontmatter}



\title{Studying Various Activation Functions and Non-IID Data for Machine Learning Model Robustness}


\author{Long Dang}
\author{Thushari Hapuarachchi}
\author{Kaiqi Xiong}
\author{Jing Lin} 

\affiliation{organization={ICNS Lab and Cyber Florida, University of South Florida},
            city={Tampa},
            state={Florida},
            country={USA}}
\begin{abstract}
\authone{Adversarial training is an effective method to improve the machine learning (ML) model robustness. Most existing studies typically consider the Rectified linear unit (ReLU) activation function and centralized training environments. In this paper, we study the ML model robustness using ten different activation functions through adversarial training in centralized environments and explore the ML model robustness in federal learning environments. In the centralized environment, we first propose an advanced adversarial training approach to improving the ML model robustness by incorporating model architecture change, soft labeling, simplified data augmentation, and varying learning rates. Then, we conduct extensive experiments on ten well-known activation functions in addition to ReLU to better understand how they impact the ML model robustness. Furthermore, we extend the proposed adversarial training approach to the federal learning environment, where both independent and identically distributed (IID) and non-IID data settings are considered. Our proposed centralized adversarial training approach achieves a natural and robust accuracy of 77.08\% and 67.96\%, respectively on CIFAR-10 against the fast gradient sign  attacks. Experiments on ten activation functions reveal ReLU usually performs best. In the federated learning environment, however, the robust accuracy decreases significantly, especially on non-IID data. To address the significant performance drop in the non-IID data case, we introduce data sharing and achieve the natural and robust accuracy of 70.09\% and 54.79\%, respectively, surpassing the CalFAT algorithm, when 40\% data sharing is used.  \authtwo{\authfour{That is,} a proper percentage of data sharing can significantly improve the ML model robustness, which is useful to some real-world applications.}}
\end{abstract}




\end{frontmatter}



\section{Introduction}
\label{introduction}

\begin{figure*}
         \centering
         \includegraphics[scale=0.70]{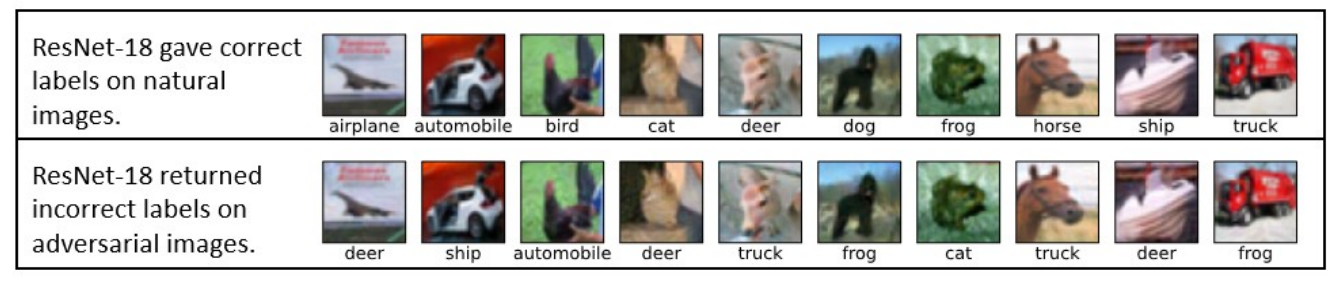}
        \caption{Illustration of $L_{\infty}$ PGD based adversarial images using a perturbation magnitude $\epsilon=\frac{8}{255}$. The first row: base images were classified correctly by a ResNet-18 network~\cite{He2015ResNet}. The second row: adversarial images generated by the PGD attacks were misclassified by the network.}
        \label{fig:adv_img} 
\end{figure*}

\authone{Adversarial training (AT) aims to defense ML models against adversarial attacks. AT involves utilizing adversarial attacks to generate adversarial examples during the training phase. Figure~\ref{fig:adv_img} illustrates clean images and their adversarial counterparts generated by an adversarial attack. With the widespread use of ML models in safety critical and cyber security systems, such as autonomous vehicles~\cite{pham2021survey}, cancer diagnosis~\cite{chaudhury2015correlation}, malware detection~\cite{DBLP:journals/access/LiXCH19}, and in-vehicle networks~\cite{Li2020AnNetworks}, adversarial attacks raise security and reliability concerns for these safety systems. Thus, it is imperative to study the ML model robustness by AT to ensure that ML models are robust against different types of adversarial attacks.}

\authone{Many defense techniques have been proposed to improve the ML model robustness against adversarial examples or adversarial attacks over the years~\cite{CHEN2023113}. They can be categorized into: (1) AT~\cite{MadryMSTV17, LinAT22, xiao2019enhancing}; (2) robustifying the model architecture by adding randomized layers~\cite{xie2017mitigating, dhillon2018stochastic,
liu2018towards, zantedeschi2017efficient} or layers with non-differentiable and discrete components~\cite{papernot2018deep}; (3) detecting adversarial examples~\cite{feiman2017detecting, lee2018simple, yang2020detect}, and (4) filtering and projecting adversarial examples~\cite{meng2017magnet}. However, randomization, detection, and filter defense techniques have been shown to be ineffective against stronger adversarial attacks~\cite{athalye2018obfuscated, carlini2017adversarial} and adaptive strategies~\cite{carlini2017magnet}. Among these defense techniques, as shown in~\cite{athalye2018obfuscated}, AT has been the only approach that is still effective against adversarial attacks. Other techniques can be incorporated into AT to further enhance the ML model robustness against complex attacks, such as parameterizing activation functions~\cite{DBLP:conf/sp/DaiMM22}, data augmentation~\cite{chen2022decision}, training a teacher model to generate soft labels~\cite{zhao2022enhanced}. While AT has demonstrated its effectiveness and flexibility in enhancing the ML model robustness, most existing AT studies limit their experiments to the Rectified linear unit (ReLU) and on centralized environments. Therefore, in this paper, we address this limitation by studying the ML model robustness using ten different activation functions and ReLU based on AT in centralized environments. We also explore the ML model robustness trained in federated learning (FL) environments.}

  \authone{In the centralized environment, we aim to adversarially train ML models to be robust against various adversarial attacks including the fast gradient sign method (FGSM)~\cite{GoodfellowFGSM}, Carlini \& Wagner (C\&W)~\cite{Carlini2017attack}, DeepFool~\cite{Moosavi-Dezfooli16}, and projected gradient descent (PGD)~\cite{MadryMSTV17} attacks at the test phase. To improve the ML model robustness against these attacks, we augment the training dataset with adversarial examples generated by an existing gradient-based iterative adversarial attack~\cite{MadryMSTV17}. We also use random noise~\cite{GuRiDenoise14}, paired with label perturbation~\cite{szegedy2016rethinking, DBLP:conf/nips/MullerKH19}. By showing the adversarial examples to the ML models during the training phase, we ensure that the models learn to correctly classify adversarial examples. We further expand the training dataset by adding random noisy examples~\cite{GuRiDenoise14}. Rather than training another ML model to provide soft labels as suggested in~\cite{papernot2016distillation}, our AT approach simply modifies the labels by using label smoothing~\cite{DBLP:conf/nips/MullerKH19}. The random noisy examples and soft labels can further enhance the ML model robustness against FGSM, C\&W, DeepFool, and PGD. Specifically, we propose an advanced AT approach to improving the ML model robustness by incorporating model architecture change, PGD based adversarial examples, Gaussian based noisy examples paired with soft labels into the iterative adversarial retraining framework in~\cite{LinAT22}. Different from ~\cite{LinAT22}, our AT approach does not consider PGD together C\&W because generating both PGD based adversarial examples and C\&W ones significantly slows the training process. The C\&W attack generates the adversarial examples with minimally disturbed perturbation by solving a constrained optimization problem instead of an unconstrained optimization problem like in PGD. Thus, we hypothesize that generating the C\&W adversarial examples is more computationally expensive than generating the PGD counterparts. However, relying on PGD for generating adversarial examples can affect the ML model robustness against the FGSM, C\&W and DeepFool attacks. To overcome this challenge, we inject small Gaussian noise into the test images. Lin et al.~\cite{LinAT22} remarked that the small Gaussian noise can disrupt the carefully crafted perturbation so that it can enhance ML model robustness against the FGSM, C\&W and DeepFool attacks while maintaining ML models' natural accuracy. The natural accuracy is the test accuracy of ML models on clean data. Therefore, our proposed centralized AT approach can improve the ML model robustness against the FGSM, C\&W, DeepFool, and PGD.} \authfour{Furthermore, we extend this centralized AT algorithm for federated AT with non-IID data with a data sharing strategy. This enables a robust training algorithm even in decentralized and heterogeneous environments, making the proposed federated AT algorithm suited for real-world federated use cases.} 
 
 \authone{ 
 Activation functions, a building block of deep neural network architecture, play a crucial role in the learning ability of deep neural networks (DNNs). Nonlinear activation functions allow DNNs to learn complex patterns in the data. In terms of improving the ML model robustness especially using AT, the exploration of activation functions remains unexplored. \authtwo{Most studies on activation functions focus on standard training~\cite{ramachandran2017searchingAF, klambauer2017self}, while a few studies examine their effectiveness in improving the ML model robustness.} Some defense methods presented the use of specialized activation functions in place of the ReLU but without using adversarial examples modified by PGD for training, like in AT~\cite{zantedeschi2017efficient, zhao2016suppressing, xiao2019enhancing}. For example, Zantedeschi et al.~\cite{zantedeschi2017efficient} proposed using bounded ReLUs to reduce the forward propagation of adversarial perturbations and crafting new training examples with Gaussian based noise, thus improving the ML model robustness against the FGSM attack. Xie, et al.~\cite{xie2020smooth} also studied improving the ML model robustness via AT~\cite{MadryMSTV17} by using ReLU\cite{nair2010rectified}, Softplus~\cite{xie2020smooth, nair2010rectified}, GELU~\cite{hendrycks2016gaussian}, and SmoothReLU~\cite{xie2020smooth}. Inspired by Xie, et al.~\cite{xie2020smooth}, we further experiment with RReLU~\cite{xu2015empirical}, SELU~\cite{klambauer2017self}, CeLU~\cite{barron2017continuously}, SiLU~\cite{hendrycks2016gaussian, ramachandran2017searchingAF}, HardSiLU~\cite{ramachandran2017searchingAF}, HardTanh~\cite{paszke2019pytorch}, TELU~\cite{fernandez2024stable}, and Mish~\cite{misra2019mish}. \authfour{Specifically, our advanced AT method in the centralized environment enables us to treat the ML model robustness as a function of model architecture, adversarial examples, and learning rate. To improve the ML model robustness against the FGSM attack~\cite{GoodfellowFGSM}, we experiment with eleven different architectures using seven smooth activation functions (Softplus~\cite{xie2020smooth, nair2010rectified}, GELU~\cite{hendrycks2016gaussian}, SiLU~\cite{hendrycks2016gaussian, ramachandran2017searchingAF}, TELU~\cite{fernandez2024stable}, CeLU~\cite{barron2017continuously}, SELU~\cite{klambauer2017self}, Mish~\cite{misra2019mish}), and four non-smooth activation functions (ReLU~\cite{nair2010rectified}, RReLU~\cite{xu2015empirical}, HardSiLU~\cite{ramachandran2017searchingAF}, and HardTanh~\cite{paszke2019pytorch}). During the training process, we adversarially train the models with three types of worst-case training examples generated by the PGD algorithm~\cite{MadryMSTV17}. We use Stochastic Gradient Descent (SGD) optimizer to iteratively update the worst-case perturbations and the model parameters. To mitigate the poor gradient quality caused by the non-smooth nature of those four non-smooth activation functions, we then consider two variants of SGD, SGD with a fixed learning rate and SGD with a linear decay learning rate schedule. In total, we collected 11 x 3 x 2 = 66 new experimental results.}
By systematically analyzing these results, we can identify which activation functions contribute the most to enhance the ML model robustness against FGSM. SmoothReLU is not included in our extensive experiments because its primary purpose is to illustrate the limitations of ReLU, which is outside the focus of our study.} Thus, our study explores the impacts of a wide range of activation functions (eleven) on improving the ML model robustness based on our proposed AT technique.

\authone{AT is known to be effective in centralized environments. However, it is still challenging to apply AT to FL environments to achieve the best ML model robustness~\cite{zizzo2020fat}. FL, known as collaborative learning, is a subfield of ML that focuses on training ML models. In FL, training data is stored on data owners' devices or \textit{clients} instead of a central storage device known as \textit{an aggregator}. Its decentralized nature leads to unique challenges such as limited communication budgets~\cite{shah2021adversarial}, inference attacks~\cite{b_IBM_FL22}, and data heterogeneity, especially the non-IID data setting~\cite{ZhaoFLNoniidData}. The non-IID data make the adoption of AT to FL more difficult because the models trained in FL converge slower than the centralized trained model. This is due to the lack of direct access to the entire dataset~\cite{b_IBM_FL22} and non-IID data with label skewness~\cite{chen2022calfat}. Recent FL studies have explored various strategies to adapt AT for FL to improve the ML model robustness, including secure aggregation~\cite{zizzo2020fat}, regularized aggregation~\cite{shah2021adversarial}, and model calibration~\cite{chen2022calfat}. Despite these efforts, there remains a performance gap when implementing AT in FL environments, particularly under the non-IID data setting, compared to the centralized environment.} \authfour{To address the challenge of non-IID data, we aim to recover the ML model robustness by extending our centralized AT approach to the FL environment. A key step in the extended federated AT algorithm is to find the best data sharing percentage that returns the highest robustness improvement. We hypothesize that the ML model robustness is dependent on the percentage of data sharing. A series of ten discrete values for the percentage parameter were selected, ranging from 0\% to 100\%, with increments of 10\%. We run the federated AT algorithm ten times with the chosen percentage values. This is done to determine which percentage of data sharing among clients will provide the best ML model robustness. Furthermore, we build a regression model to predict the ML model robustness. The trained model can predict robust test accuracies for new data sharing percentages within our selected range. Robust accuracy is the test accuracy of the trained ML model performed on the data generated by adversarial attacks. The regression model is trained using a dataset of eleven data-sharing percentages paired with robust test accuracies. These pairs are obtained from our experiments.} \authone{Thus, we first extend our proposed centralized AT approach to the FL environment, where both IID and non-IID data settings are considered.} \authfour{We create a regression model to predict the ML model robustness in non-IID data based on a new data sharing percentage.}

\authtwo{Our centralized AT approach yields significant gains in robust accuracy under the centralized environment. Compared to Lin, et al.~\cite{LinAT22}, our approach increases robust accuracy from 18.41\% to 67.96\% against FGSM and from 47\% to 83.0\% against DeepFool.}
\authtwo{Our extensive experiments on activation functions show that ReLU should be used when the learning rate is fixed. For varying learning rates, TELU is a strong alternative because it balances robust and natural accuracy with minimal trade-offs. CELU should be employed when we use strong PGD based adversarial examples for training because CeLU improves the ML model robustness and maintains generalization.}

\authtwo{We adapt our centralized AT approach to the FL environments, considering both IID and non-IID data settings. In the IID case with five clients, we achieve results slightly lower than those in the centralized case. When the number of clients increased to ten, both the robust and natural accuracies decrease further. We propose two sampling approaches: one-class and two-class non-IID to study the impacts of non-IID data on the ML model robustness. In the two-class case, both natural and robust accuracy experience significant drops compared to the IID case. The natural accuracy decreases from 66.23\% to 26.67\%. Meanwhile, the robust accuracy falls from 51.51\% to 24.54\%. \authone{To overcome the significant performance drop on the non-IID data case, we utilize an IID data sharing mechanism~\cite{ZhaoFLNoniidData}. We achieve a significant improvement in both natural and robust test accuracy, from 26.67\% to 70.09\% for the natural accuracy, and from 25.54\% to 54.79\% for the robust accuracy, respectively, when 40\% data sharing is used.}} \authfour{ While our study employs Zhao et al.~\cite{ZhaoFLNoniidData} for forming IID training sets at clients, our federated AT approach trains classifiers on adversarial rather than natural images. This ensures that the final global model is robust against adversarial attacks. Additionally, the experimental results show that our federated AT algorithm with the data sharing strategy demonstrated superior performance over the CalFAT algorithm~\cite{chen2022calfat} in both natural and robust accuracy. Our federated AT algorithm achieves a robust accuracy of 50.17\%, which marks a 43.3\% improvement over CalFAT. The federated AT model also records a better natural accuracy with 68.52\% compared to CalFAT 64.69\%.} In summary, our study makes the following key contributions.

\begin{itemize}

\item \authtwo{ We propose an advanced AT approach for improving the ML model robustness against four major adversarial attacks:  FGSM, PGD, C\&W, and DeepFool. In contrast to existing studies,  our approach includes architecture modification, soft labeling, and simplified data augmentation.}

\item \authtwo{ We perform an extensive experiment on the eleven activation functions besides ReLU, including the effect of the activation functions on ML generalization and the ML model robustness. Most activation functions can outperform ReLU in either robust accuracy or natural accuracy, but they are unable to surpass it in both robust and natural accuracy simultaneously. Hence, we conclude that ReLU is still the best choice for maintaining both natural and robust accuracy.}

\item \authtwo{We extend our centralized AT approach to the FL environment under both IID and non-IID data cases. To illustrate the non-IID data case, we include two data settings as one-class and two-class. To mitigate the reduction in the ML model robustness under the one-class and two-class settings, we utilize a data sharing mechanism.}\authfour{Through performance evaluation, we demonstrate the superiority of our federated AT approach in comparison to the CalFAT algorithm.}
\end{itemize}
\authtwo{The rest of the paper consists of the following sections. In Section~\ref{sec:background}, we provide background information on our study, followed by the research problem and associated challenges. In Section~\ref{sec:related_work}, we discuss studies related to our paper. Section~\ref{sec:method} explains our methodology. In Section~\ref{sec:evaluation}, we discuss our experimental evaluation and results. Finally, in Section~\ref{sec:conclusion}, we present the conclusions of our study along with directions for future research.  }
\section{Background and Research Problem with Challenges} \label{sec:background}
We begin with some background information about FL, adversarial attacks, centralized AT, federated AT, and \authone{activation functions}. Next, we formulate the research problem and identify the challenges addressed in this work.
\subsection{Background}
\subsubsection{Federated Learning (FL)}
\begin{figure}
    \centering
    \includegraphics[scale=0.8]{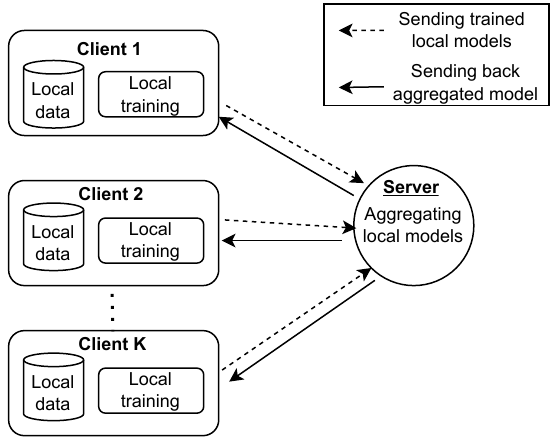}
    \caption{Model Training in FL}
    \label{fig:FL}
\end{figure}

\authtwo{Figure~\ref{fig:FL} provides an overview of a FL process. In FL, multiple clients collaborate to train an ML model while keeping their data local~\cite{b_IBM_FL22,DBLP:conf/iclr/LiHYWZ20,DBLP:journals/corr/abs-1902-04885}. First, each client trains a model using its data. Then, all the clients send their trained model parameters to the server, which aggregates the model parameters. Some widely used aggregation methods include FedAvg~\cite{DBLP:conf/iclr/LiHYWZ20}, FedCurv~\cite{DBLP:journals/corr/abs-1910-07796}, and FedProx~\cite{DBLP:conf/mlsys/LiSZSTS20}.  For example, FedAvg aggregates the models by calculating the weighted average of the model parameters~\cite{shah2021adversarial}. After aggregation, the server shares the aggregated model parameters with all the clients. The clients then train the shared model again using their data. Subsequently, they send the updated model parameters back to the server again for further aggregation. This process continues until a specified condition is satisfied, resulting in what is known as a \textit{global model}. It's important to note that all data remains local with each client during the FL process. }
\subsubsection{Adversarial Attacks}
\authone{Adversarial attacks or adversarial examples are malicious inputs to ML models, particularly neural networks by carefully crafting small, often  imperceptible perturbations~\cite{DBLP:journals/corr/SzegedyZSBEGF13}. The existence of adversarial attacks reveals a critical vulnerability in ML models and raises concerns about the security and reliability of those ML based safety critical and cyber security systems. 
Adversarial attacks can generally be categorized into targeted or untargeted attacks.
In a targeted attack, an adversary crafts an adversarial input $x^\prime$ causing the model to classify it as a chosen target class $t$, where $t$ is different from the ground truth class $y$ that $x$ belongs to. 
Conversely, in an untargeted adversarial attack, the adversary aims to generate the adversarial input $x^\prime$ that will be misclassified by the model, without targeting for any specific incorrect class. 
There are various adversarial attacks such as Fast adaptive boundary attack, Shadow attack, and Wasserstein attack~\cite{lin2021ml}. This paper primarily focuses on four popular untargeted attacks including FGSM, PGD, C\&W, and DeepFool, which are presented as follows.}


{a). \it Fast Gradient Sign Method (FGSM):}
\authone{Goodfellow et al.~\cite{GoodfellowFGSM} introduced FGSM as a one-step gradient based attack that can efficiently generate adversarial examples against DNNs. The FGSM algorithm generates perturbations ($\eta$) in the direction of the gradient of the loss function with respect to the input ($x$). That is,
\begin{equation}
    \eta = \epsilon \times \sign (\nabla_{x}\mathcal{L}(f(x),y; \theta)),
\end{equation}
where $\epsilon$ is a small, imperceptible perturbation value that determines the maximum  perturbation added to the natural (clean) inputs. Then, the adversarial image $x^\prime$ can be determined as follows:
\begin{equation}
    x^\prime = x + \eta
\end{equation}
Using the sign of the gradient of loss function ($\sign (\nabla_{x}\mathcal{L}(f(x),y; \theta))$) indicates that the algorithm moves the natural input ($x$) in the direction that most increases the loss functions. Although FGSM is widely used as a benchmark to evaluate the ML model robustness, the maximum allowed perturbation ($\epsilon$) is application specific and needs to be carefully selected to remain imperceptible to humans and undetected by adversarial detection methods. FGSM or its iterative variants is often employed in AT techniques to improve the ML model robustness~\cite{lin2020secure, Li2020AnNetworks}.}

{b). \it  Basic Iterative Method (BIM) / Projected Gradient Descent (PGD):}
\authone{BIM~\cite{DBLP:journals/corr/KurakinGB16} and PGD~\cite{MadryMSTV17} are iterative extensions of FGSM by applying FGSM multiple times. BIM typically initializes the adversarial example at the natural example while PGD begins around the natural example by adding a random perturbation  uniformly sampled from the $[-\epsilon, \epsilon]$ bound.  At each iteration, both algorithms update the point $x^t$ by:
\begin{equation}
\label{BIM}
    x^{t+1} = \Pi_{B(x_0, \epsilon)}\left(x^t + \alpha \times \sign\nabla_{x}\mathcal{L}(f(x^t),y; \theta) \right),
\end{equation}
where $t$ is the current step, $\alpha$ is the step size, and $\Pi_{B(x_0, \epsilon)}$ denotes the projection onto the $\epsilon$-ball (defined by a specific norm, often $L_\infty$ around the natural image $x_0$. The iterative nature of BIM and PGD allows them to explore a larger space of potential adversarial examples compared to the single-step FGSM, often leading to stronger adversarial examples. By exposing the models to the PGD based adversarial examples during training, the models can be more resilient to strong adversarial attacks at test time~\cite{MadryMSTV17}.} 

{c). \it Carlini and Wagner Attack (C\&W):}
\authone{The C\&W attack involves defining an attack objective function $g(x')$ such that $g(x')\leq 0$ if and only if the model $f(x')$ misclassifies the adversarial examples $x'$. Carlini et al.~\cite{Carlini2017attack} proposed to craft an adversarial example, $x'$, by iteratively solving the following optimization problem:
\begin{equation} \label{ep:C&W}
 \argmin_{x'}  \|\eta\|_p + c \cdot g(x')
\text{ such that } g(x')\leq 0
\end{equation}
where the perturbation magnitude $\|\eta = x-x'\|_p$ can be measured using the $L_2, L_0, \text{and } L_{\infty}$ norm. The $L_2$ norm attack of C\&W is among the most effective attacks to evaluate potential defenses~\cite{art2018}. $c > 0$ is a regularization parameter that allows selecting the adversarial examples that either has the smallest perturbation magnitude $\|\eta\|_p$ or has a high attack success chance. Carlini et al.~\cite{Carlini2017attack} demonstrated that the C\&W attack broke defensive distillation, a countermeasure method proposed by Papernot et al.~\cite{papernot2016distillation}, and other adversarial detection defenses.}

{d). \it DeepFool Attack:}
\authone{DeepFool attack~\cite{Moosavi-Dezfooli16} focuses on finding the shortest distance required to cross the decision boundary of the model $f(x')$, thus causing a misclassification~\cite{LinAT22}. DeepFool utilizes a linear approximation of the classifier to achieve this attack objective. Specifically, in a binary classification task, DeepFool iteratively approximates the decision boundary near a target image $x$ as a linear hyperplane. Next, DeepFool calculates the orthogonal distance from the sample $x$ to the hyperplane. By adding an overshooting pre-defined constant $\epsilon$ to the distance, the algorithm successfully finds an adversarial example $x'$, a projection of the target $x$ on the other side of this hyperplane. This process is repeatedly iteratively until the classifier's prediction $f(x')$ changes. Mathematically, the DeepFool attack can be understood as an optimization problem as follows:
\begin{equation} \label{ep:deepfool}
 \argmin_{x'} \|\eta\|_2 \text{ such that } g(x)+\nabla g(x) \cdot \eta=0,
\end{equation}
where $\|\eta\|_2$ denotes the $L_2$ distance between the natural example $x$ and the final perturbed image $x'$, and $g$ is the decision boundary. DeepFool can generate adversarial examples with a smaller perturbation compared to L-BFGS, FGSM, and Jacobian-based saliency map attacks~\cite{LinAT22}. Our AT method aims to increase the ML model robustness of deep neural networks based image classifiers against the FGSM, C\&W, DeepFool, and PGD attacks.}

\subsubsection{Adversarial Training (AT)}

This subsection reviews centralized and federated AT relevant to our study.

{a). \it Centralized AT:}
\authone{
Despite the high performance of deep neural networks (DNNs) on various tasks, they are vulnerable to adversarial examples. These maliciously perturbed inputs can cause DNNs to make incorrect predictions, highlighting significant security concerns. Madry et al. \cite{MadryMSTV17} introduced AT as one of the most effective defense techniques to improve the model robustness of ML models against strong adversarial attacks. 
AT involves training a ML model on a dataset that includes both natural (clean) and adversarial examples. The goal is to train a model that is robust to adversarial perturbations. Centralized AT refers to the traditional setting which the entire training dataset is stored in one place. Specifically, in each iteration, adversarial examples are generated from the current training data using common attack methods such as FGSM, PGD, C\&W, and DeepFool~\cite{MadryMSTV17, LinAT22}. Then, the model is trained using an augmented training dataset with these adversarial examples, together with the natural examples. The trained model will learn to correctly classify the adversarial examples. Hence, the AT process can be formulated as a min-max optimization problem~\cite{MadryMSTV17, shah2021adversarial}:
\begin{equation}
    \label{eq:AT}
    \argmin_{\theta}  \mathbb{E}_{(x, y) \sim D} \left[\argmax_{\|\delta\|_p \leq \epsilon} \mathcal{L}(f(x+\delta),y;\theta)\right],
\end{equation} 
where $D$ be a training set comprising pairs of samples $(x \in \mathbb{R}^n, y \in k)$; $k$ is a set of all possible classes given in the data. $\mathcal{L}$ denotes a loss function for a deep neural network $f$ with model weights $\theta \in \mathbb{R}^p$ where $p$ is an integer representing the dimension of the model weights. $\delta$ represents the perturbation added to the natural example $x$. $\epsilon > 0$ denotes the adversarial perturbation radius, which is the maximum perturbation allowed.} 

\authone{While the AT process provides a promising solution to improve the ML model robustness, its effectiveness is influenced by the choice of activation functions; e.g., ReLU, GELU, Softplus, ELU, and SILU were studied for AT in~\cite{xie2020smooth}. Using an activation function in AT to improve the ML model robustness is not an easy task. For instance, ReLu is one of the most popular activation functions to be used in the AT process. However, using ReLU in AT presents two key challenges: dead neurons~\cite{maas2013rectifier}, and its non-smooth nature~\cite{xie2020smooth}. These challenges hinder the optimization in Equation~\ref{eq:AT}~\cite{maas2013rectifier,xie2020smooth}, ultimately reducing the ML model robustness. Thus, these challenges underscore the importance of exploring alternative activation functions to enhance the ML model robustness against adversarial attacks.}

{b). \it Federated AT:}
\authtwo{In federated AT, the clients adversarially train their local models rather than performing standard training before sending the model parameters to the aggregator. Specifically, each client solves the optimization problem defined in  Equation~\ref{eq:AT} using its own dataset ($D_k)$ to conduct local AT. The server then aggregates the adversarially trained model parameters in each round using a fusion function $F$.  Suppose there are  $K$ clients, and $\theta_i^t$ denote the model trained adversarially by client $i$ for a given round $t$. The aggregation of these models across clients is expressed mathematically in Equation \ref{ep:fl_fusion}.}
\begin{equation} 
\label{ep:fl_fusion}
\theta^{t+1} = F(\theta_{1}^{t}, \theta_{2}^{t},...,\theta_{K}^{t}) 
\end{equation}
Because each client's local data is not shared with other clients, FL must \authone{address the challenges arising from the non-IID data. When AT is incorporated into FL, the non-IID data can create divergent ML models, thus yielding a suboptimal robust global model. Thus, addressing the issue of non-IID data requires innovative approaches to improve the ML model robustness in FL environments.} 
\subsubsection{Activation functions}
\authone{In this work, we extensively examine the impact of the ten activation functions, in addition to ReLU, on the ML model robustness based on our AT technique. To provide context for this analysis, we summarize the basic characteristics of these functions in this section. For a survey of activation functions in ML, we refer the readers to the comprehensive work by Dubey et al.~\cite{dubey2022activation}.} 
\authone{a). \it Rectified linear unit (ReLU)~\cite{nair2010rectified}:}
\begin{equation}
\authone{
\textbf{ReLU}(x) =
    \begin{cases}
        x & \text{if } x \geq 0 \\
        0 & \text{ otherwise }
    \end{cases}
}
\end{equation}
\authone{ReLU is more computationally efficient than Logistic, Sigmoid, and Tanh because the ReLU zeroizes all the negative inputs. On the other hand, because only the positive input values are kept, it limits the neural network's learning capability. Furthermore, its gradient is also zero for the inputs less than zero, which creates the dying ReLU problem~\cite{lippe2024uvadlc} during the training phase.}
\authone{{b). \it Randomized leaky rectified linear unit (RReLU)~\cite{xu2015empirical}:}}
\begin{equation}
\authone{
\textbf{RReLU}(x) =
    \begin{cases}
        x & \text{if } x \geq 0 \\
        a \times x & \text{ otherwise }
        ,
    \end{cases}
}
\end{equation}
\authone{where $a$ is a random variable sampled from a uniform distribution
$\mathcal{U}(\text{l}, \text{u})$. The RReLU's gradient is random and non-zero for the negative inputs. By making this change, RReLU enables backpropagation and eliminates the dying ReLU problem. We will set the parameters $\text{l} = \frac{1}{8}$ and $\text{u} = \frac{1}{3}$ for our training phase. These values were empirically determined in~\cite{xu2015empirical}. To make the prediction consistent for the negative values, during the testing phase, we will deterministically fix $a$ by setting $a = \frac{\text{l} + \text{u}}{2}$.}
\authone{{c). \it Scaled exponential linear unit (SELU)~\cite{klambauer2017self}:}}
\begin{equation}
\authone{
\textbf{SELU}(\alpha, \lambda, x) = 
    \lambda \times \begin{cases}
         x & \text{if } x \geq 0 \\
        \alpha \times (e^{x} - 1) & \text{ otherwise ,}
    \end{cases}
    }
\end{equation}
\authone{with the predefined constant $\alpha \approx$ 1.6733 and $\lambda$ $\approx$ 1.0507 as specified in~\cite{klambauer2017self}. The SELU was first used in a feedforward neural network called self-normalizing neural network (SNN)~\cite{klambauer2017self}. As shown in ~\cite{klambauer2017self}, SELU SNNs and a weight initialization strategy were better than traditional ReLU and RReLU feedforward networks in 121 data sets from the UCI ML repository.}
    
\authone{{d). \it Continuously differentiable exponential linear unit (CeLU)~\cite{barron2017continuously}:}} 
\begin{equation}
\authone{
\textbf{CeLU}(\alpha, x) =  
    \begin{cases}
        x & \text{if } x \geq 0 \\
        \alpha \times (e^{x/\alpha} - 1) & \text{ otherwise },
    \end{cases}
    }
\end{equation}, \authone{with the shape parameter $\alpha$'s is learnable. To tackle ReLU's non-smooth nature, the CeLU uses the exponential function to transform the negative input values. CeLU can be a strong alternative for ReLU in AT because of the following reasons: first, computing CeLU over all positive input values are the same as ReLU. Second, CeLU avoids the dying ReLU problem by introducing the exponential function for all negative inputs. Third, by treating the parameter $\alpha$ learnable, Barron et al.~\cite{barron2017continuously} automatically tuned $\alpha$ for different datasets. In our experiments, CeLU outperformed ReLU on both natural and robust accuracy when we used PGD with a high number of iterations to generate adversarial examples. }
\authone{{f). \it Sigmoid-weighted linear unit (SiLU)~\cite{hendrycks2016gaussian, ramachandran2017searchingAF}:}}
\begin{equation}
\authone{
\textbf{SiLU}(x) = x \times \text{Sigmoid}(x),
}
\end{equation}
\authone{where the Sigmoid activation function is defined as $\text{sigmoid}(u) = 1 / (1 + e^{-u})$. The SiLU is an improvement of the Sigmoid function, which suffers from the vanishing gradient problem~\cite{hendrycks2016gaussian, ramachandran2017searchingAF}. SiLU, first mentioned in ~\cite{hendrycks2016gaussian}, is also known as Swish in~\cite{ramachandran2017searchingAF}. Ramachandran et al. ~\cite{ramachandran2017searchingAF} used automatic search techniques to discover SiLU. As \authtwo{shown} in~\cite{ramachandran2017searchingAF}, the SiLU, used by the preactivation ResNet-164~\cite{he2016identity} and Wide ResNet 28-10~\cite{zagoruyko2016wide}, outperform\authtwo{s} the ReLU for image classification on CIFAR-10 and CIFAR-100 datasets. In our experiments, we will use ResNet-18~\cite{He2015ResNet}, a smaller version of the ResNet model family, and we hypothesize that ResNet-18 with SiLU can match or outperform the same model with ReLU in improving robust accuracy and possibly maintaining natural accuracy using our centralized AT.} 
\authone{{g). \it HardSiLU~\cite{ramachandran2017searchingAF}}}:
\begin{equation}
\authone{
\textbf{HardSiLU}(x) =
\begin{cases}
    0 & \text{if } x \leq -3, \\
    x \times \frac{x + 3}{6} & \text{if } -3 < x < 3 \\
    x & \text{if } x \geq 3
\end{cases}.
}
\end{equation}
\authone{HardSiLU~\cite{ramachandran2017searchingAF} can be considered as a combination of ReLU and SiLU for computational efficiency. Unlike SiLU, HardSiLU uses a quadratic function for inputs ranging from -3 to 3 rather than the computationally expensive exponential function. SiLU adopts ReLU's behaviors for the input values less than -3 and more than 3. Thus, HardSiLU significantly reduces computation costs when conducting the ML training and test phases on mobile devices with limited resources~\cite{ramachandran2017searchingAF}. Even though HardSiLU is designed to be efficient, we aim to evaluate whether HardSiLU can serve as an effective alternative to ReLU in ResNet-18 for centralized AT, focusing on its impact on ML robustnes.}
\authone{{h). \it Hard hyperbolic tangent (HardTanh)~\cite{paszke2019pytorch}}}:
\begin{equation}
\authone{
\textbf{HardTanh}(x) = 
        \begin{cases}
            min\_value & \text{ if } x < min\_value \\
            max\_value & \text{ if } x > max\_value \\
            x & \text{otherwise} \\
        \end{cases}.
        }
\end{equation}
\authone{ We know that the Tanh function is defined as: 
$\text{Tanh}(x) = \frac{\exp(x) - \exp(-x)}{\exp(x) + \exp(-x)}.$ In HardTanh, for both large negative and positive values, there is no computation, and the outputs are set to $min\textunderscore value$ and $max\textunderscore value$, respectively. Like ReLU, the function returns the input value when an input is between $min\textunderscore value$ and $max\textunderscore value$ instead of computing the exponential functions. Thus, HardTanh is more computationally efficient than the traditional Tanh function. We investigate its potential impact on improving the ML model robustness when compared to ReLU. In our experiments, we will use the default values for $min\textunderscore value = -1$ and $max\textunderscore value = 1$ provided in PyTorch~\cite{paszke2019pytorch}. }
\authone{{i). \it Gaussian error linear unit (GELU)~\cite{hendrycks2016gaussian}}}:
\begin{equation}
\authone{
\textbf{GELU(x)}=\left(\frac{x}{2}\right)\left[1+\tanh \left(\sqrt{\frac{2}{\pi}}\left(x+ c x^3\right)\right)\right]}
\end{equation}, with a constant $c = \geluconst$.
\authone{As shown in~\cite{hendrycks2016gaussian}, a 9-layer convolutional neural network with the GELU consistently has lower natural accuracy than the ReLU ones. The model's parameters were optimized with \authtwo{the Adam optimizer for 200 epochs. Furthermore, rather than maintaining a fixed learning rate throughout the training process, the Adam optimizer used a linear decay learning rate schedule with linear decay over time.}}

\authone{{k). \it Stable parametric softplus (Softplus)~\cite{xie2020smooth, nair2010rectified}:}}
\begin{equation}
\authone{
\textbf{Softplus}(\beta, x) = \frac{1}{\beta} \times \log(1 + e^{\beta \times x}).
}
\end{equation}
\authone{Softplus is a smooth approximation of ReLU. In~\cite{xie2020smooth}, the ResNet-50 model with the Softplus ($\beta$ = 10) has comparable natural accuracy and better robust accuracy than the ReLU based corresponding model on CIFAR-10. In our experiments, we will set $\beta $ to 1 as suggested in~\cite{paszke2019pytorch}.}
\authone{{j). \it Hyperbolic tangent exponential linear unit (TELU)~\cite{fernandez2024stable}:}}
\authone{
\begin{equation}
    \textbf{TELU}(x) = x\times \tanh(e^{x}).
\end{equation}}
\authone{TELU is a recently proposed activation function by Fernandez et al.~\cite{fernandez2024stable}. TELU~\cite{fernandez2024stable} combines exponential and hyperbolic tangent functions to introduce non-linearity while preserving smooth gradients. In standard training, the authors demonstrated that Resnet-50 models with TELU achieve stable training performance across various optimizers, including SGD, SGD with momentum, RMSprop, and Adam. In this paper, we extend the study of TELU by evaluating its effectiveness on improving the ML model robustness via our proposed AT technique.}
\authone{{h). \it Mish~\cite{misra2019mish}:}}
\authone{
\begin{equation}
\textbf{Mish}(x) = x \times \tanh(\text{Softplus}(x)).
\end{equation}}
\authone{Unlike ReLU, Mish is \authtwo{a} smooth and non-monotonic activation function, which  avoids the dying ReLU problem~\cite{misra2019mish}. Misra et al.~\cite{misra2019mish} show\authtwo{ed} that Mish increases the natural accuracy over ReLU and Swish for image classification on the CIFAR-100 dataset.}
\subsection{Research Problem with Challenges}
\authone{With an increasing and widespread use of ML especially DNNs across numerous critical real world applications, ensuring the the ML model robustness against worse-case noises, adversarial examples, and their operational issues is crucial. Despite the outstanding performance of DNNs on various complex tasks, they are inherently vulnerable to adversarial attacks. In this study, we focus on the adversarial attacks under white-box settings, where the adversary is assumed to have complete knowledge of trained models, including their architecture and parameters. AT has been one of the most effective defense techniques against adversarial attacks under centralized environments. 
Existing AT approaches mostly \authone{employ} ReLU as the activation function and rarely explore other advanced activation functions~\cite{xie2020smooth}. To address this research gap, we perform a comprehensive study on \authone{how} the ten activation functions beside ReLU \authone{make impacts} on the ML model robustness.} \authtwo{Our analysis reveals three key challenges. First, selecting the best activation function is difficult due to a performance imbalance between natural and robust accuracy—some functions surpass ReLU in natural accuracy but underperform in robust accuracy. Second, the computational complexity of certain activation functions increases training time, slowing overall execution~\cite{fernandez2024stable}. Third, some activation functions can lead to slow convergence, a problem further exacerbated by adversarial training, which introduces additional optimization difficulties~\cite{xie2020smooth}.}
\authone{When it comes to directly applying centralized methods to a decentralized environment, significant challenges include performance degradation due to non-IID data~\cite{zizzo2020fat}, increased convergence time~\cite{shah2021adversarial}, and new attack surfaces~\cite{b_IBM_FL22}. Therefore, existing AT require further development and adaptation to effectively address the unique challenges posed by decentralized environments especially FL, particularly with non-IID data. The primary goal of the research presented in the paper is to tackle the non-IID data challenge by improving existing AT techniques and extending them to FL with a strong focus on the impact of IID and non-IID data. The research also aims to mitigate the performance gap observed between centralized and federated AT.}
\section{Related Work}\label{sec:related_work}
\authtwo{In this section, we summarize prior research \authone{according to the following categories: } centralized AT, activation functions for the ML model robustness, and federated AT.}
\textbf{Centralized AT:} \authtwo{Madry et al.~\cite{MadryMSTV17} proposed adversarial training as a powerful remedy to improve the ML model robustness against adversarial attacks. Later, Lin, et al.~\cite{LinAT22} introduced an advanced adversarial retraining method for further enhancing the ML model robustness. Their approach incorporated several techniques, including label smoothing~\cite{DBLP:conf/nips/MullerKH19}. Specifically, Lin, et al.~\cite{LinAT22} assigns a dominant probability (e.g., 0.8) to a target value of '1' for the true class, and uniformly distributes the remaining probability (e.g., 0.2) among incorrect classes to the target value of '0'. Additionally, the assigned probabilities varied based on the perturbation intensity ($\epsilon$) of adversarial examples. In comparison, our approach simplified this process by applying the same probability scheme to all adversarial examples. Furthermore, while Lin et al.~\cite{LinAT22} utilized adversarial examples generated from both C\&W and PGD attacks during training, we rely solely on PGD-generated examples, thus significantly reducing the overhead computational costs. We also modified the official ResNet-18 architecture~\cite{He2015ResNet} to enhance the ML model robustness against adversarial attacks. }
\authtwo{\textbf{Activation functions for the ML model robustness:} Most of the \authone{existing} ML studies use ReLU as the activation function due to its effectiveness and simplicity~\cite{DBLP:journals/jmlr/Zhang0Z24}. Even though ReLU is possibly the best option for standard training, recent studies demonstrated that smooth non-parametric activation functions~\cite{xie2020smooth} and smooth parametric activation functions~\cite{DBLP:conf/sp/DaiMM22} can surpass ReLU on improving robust accuracy in AT. Dai et al.~\cite{DBLP:conf/sp/DaiMM22} proposed improving the ML model robustness by studying learnable parametric activation functions in conjunction with AT. They selected ReLU, Softplus, and SiLU as the starting points and demonstrated that the smooth forms of these functions obtained by parameterizing and hyper-parameter tuning can significantly improve the ML model robustness over ReLU.   
Moreover, the non-parametric smooth forms of ReLU, Softplus, Silu, GELU, and Elu in the AT process can improve the ML model robustness with no increase in computational cost~\cite{xie2020smooth}.
In contrast, we investigated seven smooth activation functions (Softplus, GELU, SiLU, TELU, CeLU, SELU, and Mish) and four non-smooth activation functions (ReLU, RReLU, HardSiLU, and HardTanh), and evaluated their impact on the ML model robustness via AT. } 
\authone{\authtwo{\textbf{Federated AT:}}
\authtwo{Ludwig et al.}~\cite{b_IBM_FL22} grouped potential
adversaries against FL into two classes: insiders or outsiders. Insider adversaries can be the aggregators and the clients of FL. All other potential adversaries are considered outsiders, for example, the users of the text messaging application who are not the aggregators and the clients. Outsider attacks contain two main types\authtwo{:} training-time attacks and test-time attacks. In test-time attacks, an outsider adds a small perturbation on test data to make untargetted/targeted misclassifications 
without tampering with the FL training process. Zizzo et al.~\cite{zizzo2020fat} introduced federated AT (FAT) to improve the ML model robustness against adversarial outsiders. Shah et al.~\cite{shah2021adversarial} proposed an aggregation function called FedDynAT for AT when the client's training data were non-IID and the communication rounds were constrained. However, both studies~\cite{zizzo2020fat, shah2021adversarial} observed a significant drop in the natural accuracy and robust accuracy as opposed to centralized training. Furthermore, Dang et al.~\cite{dang2023improving} modified the ResNet-18 model's architecture to enhance the robust accuracy under both centralized AT and FAT against the FGSM attacks.} \authfour{Another important work is CalFAT~\cite{chen2022calfat}, a calibrated federated AT algorithm. Chen et al.~\cite{chen2022calfat} addressed the challenge of non-IID data with label skew when conducting adversarial training in FL. To overcome the non-IID data challenge, CalFAT adaptively calibrated logits to balance class distributions. This calibration then improved model convergence and the ML model robustness. Although both methods aimed to improve the ML model robustness in the non-IID data case,  \authtwo{our} federated  AT \authtwo{approach} solved a broader non-IID data issue mainly through data sharing. This study compares the natural and robust accuracy of the proposed federated AT with the CalFAT to assess its broader effectiveness.}
\section{Methodology} \label{sec:method} 
\begin{figure}[!ht]
		         \centering
		         \includegraphics[scale=0.50]{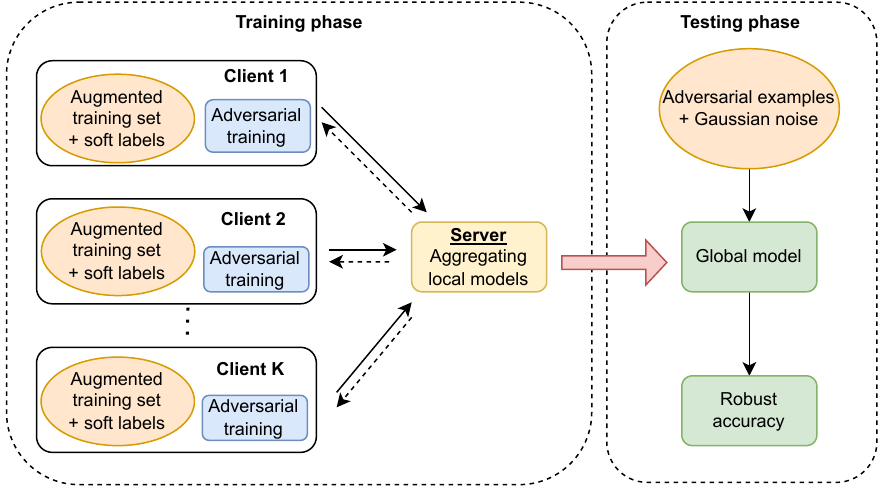}
		        \caption{Federated AT procedure.}
		        \label{fig:process}
\end{figure}
In this section, we present our proposed federated AT process. The process begins with a federated AT phase followed by a test phase. In the federated AT phase, we execute Algorithm~\ref{algo:fl} to collaboratively train a robust global model and the training algorithm can support various activation functions without any modification to the model architecture. In the test phase, we evaluate the ML model robustness of the global model against various adversarial attacks and noise.\authone{Figure~\ref{fig:process} illustrates the proposed federated AT approach. In this process, each client augments its local data with adversarial examples (e.g., generated using PGD and Gaussian noise examples) and uses soft labels instead of the hard labels given in the training dataset. After training the local models, the server then aggregates those local models to form a robust global model. The final global model then can be tested for the ML model robustness against adversarial attacks.} We further discuss the key components of our approach below.
\subsection{Federated AT phase}
\subsubsection{Initialization}\label{sec:data_sharing}
\authone{
Non-IID data poses a significant challenge in the training of high-performance FL models. 
In particular, the non-IID data from each client can hinder the convergence of the global model and decrease both its natural and robust accuracy, especially when extending AT to the federated learning environments as highlighted in~\cite{ZhaoFLNoniidData, shah2021adversarial}. 
This subsection aims to outline the methodology used to address the non-IID data problem within the federated AT phase. To mitigate the non-IID data problem, we utilized a data sharing strategy proposed by Zhao et al.~\cite{ZhaoFLNoniidData} as the primary technique. 
By providing a globally shared IID dataset containing examples from each class to FL clients, the local data distribution becomes less skewed toward certain classes~\cite{ZhaoFLNoniidData}. 
To create the globally shared dataset, we first sorted the training dataset by class labels and then processed each class individually. For each class, we randomly selected 1,000 samples without replacement. We combine all the selected samples to form the global shared dataset. This ensures that the global shared dataset maintains a uniform class distribution. The remaining training data was partitioned among the clients to create the non-IID local datasets. Zhao et al.~\cite{ZhaoFLNoniidData} also introduced a parameter $\alpha$ for controlling the different levels of data sharing across FL clients. By varying $\alpha$ from 0 to 1 in increments of 0.1, the proportion of shared images increases incrementally, starting with no shared data ($\alpha$ = 0) and reaching using the full global shared dataset ($\alpha$ = 1). We explore the impact of $\alpha$ on robust accuracy in Section~\ref{Discussion}.} 
While designing our federated AT algorithm, we want our procedure to account for the following two cases: 1) all clients have IID data, and 2) all clients do not have IID (non-IID) data. In the IID case, because each client is expected to have the same amount of data with the same class distribution, the data sharing method is not integrated in this case. In the non-IID case, all the clients have the same amount of data with non-uniform class distributions. We adopted the two settings described in Zhao et al.~\cite{ZhaoFLNoniidData}. In the first setting, we considered the worst non-IID data, where each client has data from only one class. In the second setting, we randomly assigned data from two classes to each client.\authtwo{Figure~\ref{fig:non_iid} shows the non-IID data distribution among ten clients under the one-class and two-class settings.}
\authone{In our experiments, when we set $\alpha$ = 0.5, we randomly sampled 500 data points per class from the global shared dataset. That means each client received either its one-class or two-class non-IID data partition in addition to a random sample of 500 data points for each class given by the globally shared dataset. Each client then performed the local AT on a combined training dataset comprising of its local non-IID data and the shared IID data. In Section~\ref{FAT_Non_IID}, we demonstrated the effectiveness of the data sharing approach in increasing both natural and robust accuracy compared to the same non-IID data settings without data sharing.} Next, we generated adversarial examples using PGD and the Gaussian distribution on the combined local dataset.
\subsubsection{Crafting adversarial examples}
\begin{figure}
         \centering
         \includegraphics[scale=0.5]{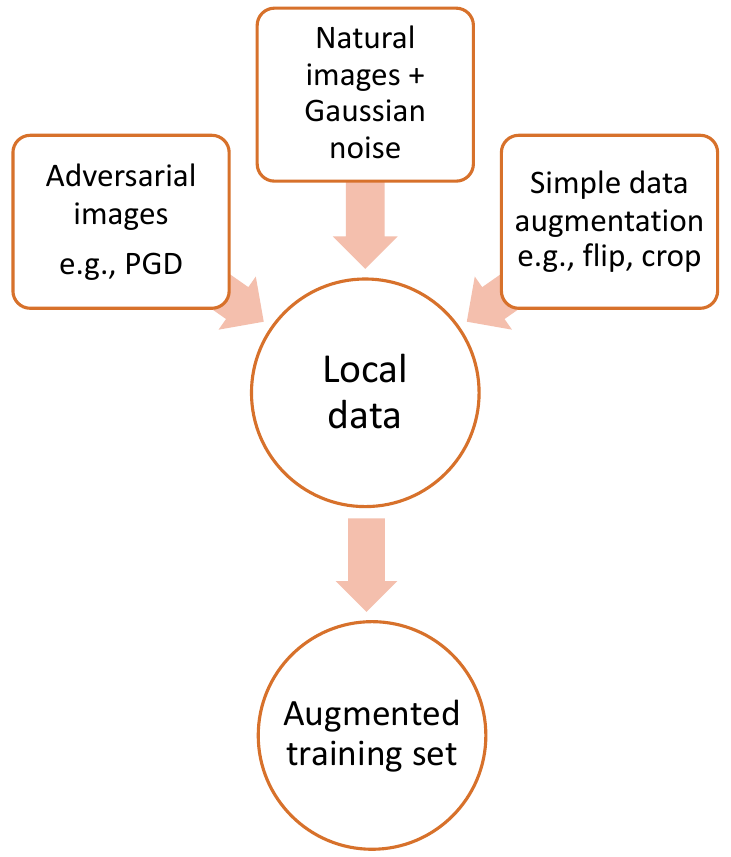}
        \caption{Augmenting training set for AT by each client. This figure details the data augmentation process performed by each client in the federated AT method.}
        \label{fig:data_aug}
\end{figure}

\authone{We augmented the local data in each client, as shown in Figure~\ref{fig:data_aug}. We first expanded the training set with adversarial examples crafted from PGD. We choose the costly PGD over the fast FGSM because the model trained on PGD based adversarial examples has better robust accuracy against FGSM, C\&W, and DeepFool attacks than the same model trained with FGSM based adversarial examples. We also limited the maximum perturbation to be unrecognizable by human eyes so that the model could learn to recognize adversarial images with minimal perturbations like those generated by C\&W and DeepFool attacks. It is also important for the model to be robust against random noise. We simulated the noise using Gaussian probability distribution and we added the Gaussian noise to natural images. Furthermore, we also applied simple data augmentation methods like horizontal flipping and random cropping with padding to the training set to reduce overfitting ~\cite{DBLP:journals/jbd/ShortenK19}.} 
\subsubsection{Soft labels}
\authone{Soft labels were used during the federated AT phase. By combining crafting adversarial examples and soft labels, we aim to enhance the global model’s robust test accuracy against unseen adversarial attacks including FGSM, C\&W, and DeepFool attacks than the federated AT with hard labels. 
The soft label $\mathbf{y}^{SL}$ for an input belonging to class $c$, where $c$ is an integer between 1 and the total number of classes $N$ is defined as follows.} 
\begin{equation}
\label{eq:soft_label}
    y_i^{SL}= \begin{cases}
       1- \frac{N-1}{N}\alpha & \text{if } i =c\\
         \frac{\alpha}{N} & \text{if } i \neq c,
    \end{cases}
\end{equation}
where $\alpha$ is a label smoothing parameter that is close to 0. \authone{This approach assigns a high value close to 1 to the true label and distributes a near zero value to the false labels, ensuring that the sum of all the values are 1. 
Lin et al.~\cite{LinAT22} applied a soft labeling approach to AT where the values assigned to the labels dependent on the types of the training samples and the perturbation limit of adversarial examples. In our study, we simplified this by applying Equation~\ref{eq:soft_label} to derive the soft labels without any conditions on clean examples, Gaussian noise examples, and PGD-based adversarial examples.} 
\subsubsection{Activation functions}
\authone{During the training phase, the choice of activation functions is critical because it affects the performance of AT~\cite{xie2020smooth}. ReLU is a commonly used activation function in improving the ML model robustness through AT. Xie et al.~\cite{xie2020smooth} proposed replacing ReLU with modern activation functions, especially ReLU's smooth approximations. \authtwo{When} there are many new functions introduced to substitute ReLU, we carefully choose those ten different activations that have been widely used in existing literature and study \authtwo{how they impact} the ML model robustness through AT. \authtwo{That is,} we conduct\authtwo{ed} an extensive experimental study on \authtwo{those} eleven different activation functions including the ReLU in a centralized environment. They are ReLU~\cite{nair2010rectified}, RReLU~\cite{xu2015empirical}, Softplus~\cite{xie2020smooth, nair2010rectified}, GELU~\cite{hendrycks2016gaussian}, SELU~\cite{klambauer2017self}, CeLU~\cite{barron2017continuously}, SiLU~\cite{hendrycks2016gaussian, ramachandran2017searchingAF}, HardSiLU~\cite{ramachandran2017searchingAF}, HardTanh~\cite{paszke2019pytorch}, TELU~\cite{fernandez2024stable}, and Mish~\cite{misra2019mish}. Thus, we first constructed eleven different ResNet-18 model by varying the eleven activation functions in each hidden layer while keeping all other network parts exactly the same. We then trained these models using our centralized AT method and measure the ML model robustness using \authtwo{the} FGSM attack. Finally, we compared the robust accuracies obtained by the eleven models to find the activation function with the highest robust accuracy. \authtwo{Based on our experimental results in  Section~\ref{variants_of_ReLU__AF}, ReLU was the best \authone{choice} for the activation function to maintain both robust and natural accuracy. Therefore, we chose ReLU as the activation function in the FL environment.}}
\subsubsection{Local learning rate}\label{sec.lr}
\authtwo{Replacing ReLU with the advanced activation functions improves the gradient quality but can result in complex loss landscapes with both down-hill, up-hill, and flatten areas. In the early epochs, the SGD can effectively guide the training process to go down-hill by following the direction with the steepest slope. However, in later epochs, when the loss landscape becomes flat, SGD struggles to find a good direction to continue going down-hill by using a fixed learning rate. We hypothesized that \authone{in order for us} to find a good local optimal, in the early epochs, SGD needs a high learning rate\authone{; conversely,} in later epochs SGD requires a small learning rate. Based on our hypothesis, we used a piece-wise approach to gradually reducing the learning rate. We used a single linear function with a slope of 1/10 to obtain the learning rate based on the current epoch. Eq.~\ref{eq.lr} defines this approach where $\eta$ denotes the current learning rate\authone{,} and $m$ denotes the current local epoch. Specifically, we set the initial learning rate to 0.001 when the current epoch is less than 100. If the current epoch is 100 or greater, we divided the learning rate by ten. If the current epoch is greater than or equal to 150, the learning rate is divided by ten again.}
\begin{equation}
\authtwo{
    \eta= \begin{cases}
        0.001 & \text{if } m < 100\\
        0.0001 & \text{if } 100 \le m < 150 \\
        0.00001 & \text{if } 150 \le m\\
    \end{cases}}
    \label{eq.lr}
\end{equation}
\subsubsection{Global model update}
\authone{Updating the global model's parameters in federated AT involves aggregating the locally trained models from all participating clients. This iterative process ensures that the global model incorporates knowledge from all clients. Algorithm \ref{algo:fl} provides a detailed representation of our federated AT approach. In each communication round $t$ between clients and server, each client $k$ iteratively trains his/her local model using his/her dataset comprising of a combination of local and a portion of the globally shared data for $E$ epochs. Then, the client sends the weights of their adversarially trained model $\theta_{k}^{t}$ to the server. The server aggregates these weights using a fusion function $F$. In our study, the global model weights $\theta^{t+1}$ are updated using Federated Average (FedAvg)~\cite{DBLP:conf/aistats/McMahanMRHA17}. The FedAvg computes a weighted average of all local models based on their data contribution ratio $\frac{|D_k|}{|D|}$, as shown in the following equation.  
\begin{equation}\label{fedavg}
    \theta^{t+1} = \frac{1}{|D|} \sum_{k=1}^K{|D_k| \theta_{k}^{t}},
\end{equation} where $|D_k|$ is the size of the dataset belongs to the client $k$ and $|D|=\sum_{k=1}^K |D_k|$.
The server sends back the updated weights $\theta^{t+1}$ to the clients. This completes one communication round between the server and clients. The process is repeated for $R$ communication rounds before obtaining the final global model. Our study shares a similar context with Shah et al.~\cite{shah2021adversarial}. While Shah et al.~\cite{shah2021adversarial} focus on improving the effectiveness of AT with non-IID data in a constrained communication budget, we study on improving the ML model robustness via AT with non-IID data against various unseen and strong adversarial attacks and random noise.} 

\begin{figure*}[!t]
\centering
\begin{minipage}{0.95\textwidth}
\begin{algorithm}[H]
\begin{algorithmic}[1]
\STATE \textbf{Input} $K$: number of clients; $R$: number of communication rounds; $\mathcal{D}_k$: the training set for client $k$; $\mathcal{G}$: globally shared training set (created in the initialization phase); $\alpha$: the fraction of $\mathcal{G}$ distributed to each client; $A_T$: augmented training set; $y$: the training set labels; $\alpha_{SL}$: a label smoothing parameter; $E$: number of local training epochs per round; $\theta^{0}$: initial model weights.
\STATE \textbf{Output}: The global model that is robust to adversarial attacks. 
    \STATE \textbf{Initialization}:
        \STATE Create $\mathcal{G}^{IID}$ with an uniform class-wise distribution by randomly choosing $N$ instances per class from the training dataset
        \STATE Create $\mathcal{D}^{non-IID}_k$ with an non-IID class-wise distribution by randomly choosing the remaining instances per class from the training dataset
        \STATE Create $\mathcal{S}^{IID}$, a random $\alpha$ portion of $\mathcal{G}$
        \STATE Combine $\mathcal{S}^{IID}$ and $\mathcal{D}^{non-IID}_k$ to form a clean training dataset $\mathcal{D}_k$ for each client $k$
    \FORALL{ $t = 1$ \textbf{to} $R$}
        \STATE Aggregator sends the model weight $\theta^{t}$ to all clients
        \FORALL{$k = 1$ \textbf{to} $K$}
            \STATE Apply PGD, Gaussian noise, and data augmentation on $\mathcal{D}_k$ to create the augmented training set $A_{T}$ 
            \STATE Calculate $\mathbf{y}_{SL}$ based on Equation~\ref{eq:soft_label}
            \STATE $\theta_{k}^{t+1} = AT(E, \theta^{t}, A_{T}, \mathbf{y}_{SL})$; 
            \STATE Send $\theta_{k}^{t+1}$ back to the aggregator;
        \ENDFOR
        \STATE $\theta^{t+1} = FedAvg(\theta_{1}^{t}, \theta_{2}^{t},...,\theta_{K}^{t})$; 
    \ENDFOR
\end{algorithmic}
\caption{Federated AT with non-IID data.} 
\label{algo:fl}
\end{algorithm}
\end{minipage}
\end{figure*}
\subsection{Adversarial testing phase}
\authone{Our testing phase was designed to be robust against various adversarial examples. Specifically, we evaluated the ML model robustness on adversarial examples generated by the FGSM, C\&W, DeepFool, and PGD attacks. During this phase, our aim is to improve the ML model robustness by injecting a small Gaussian noise to adversarial examples. The added noise can distort the deliberately calculated adversarial perturbations~\cite{LinAT22, DBLP:conf/icc/LinNX20}. Because the ML models are trained with Gaussian noise based examples, the robust accuracy can be improved without affecting the natural accuracy. Thus, the models can be more robust against adversarial attacks than in the standard testing phase without adding Gaussian noise.}
\section{Experimental Evaluation and Results} \label{sec:evaluation}
This section presents the implementation and evaluation of our centralized AT and federated AT approaches for both IID and non-IID data. We assess the efficacy of the federated AT approach by comparing the natural and robust accuracy to that of the centralized AT ones.
\subsection{Software and Hardware}
\authone{Our study utilized Python 3.7.6 and PyTorch 1.13.1 to generate adversarial examples for both training and testing. For hardware resources, we primarily used an NVIDIA A40 GPU card to accelerate computations, and when the A40 was occupied by other university researchers, we used an NVIDIA GeForce GTX 1080 Ti card.} While the Adversarial Robustness Toolbox (ART)~\cite{art2018} used in the study~\cite{LinAT22} offers a standardized and automated process for AT and the ML model robustness evaluation, our approach allows for more granular control over generating the adversarial examples. Despite these differences in implementation, both studies aim to evaluate the ML model robustness. \authone{It is worth mentioning that our implementation of FL is a simulated version based on a Github code base~\cite{AshwinRJ_FL_PyTorch}. The FL program does not support for parallelization of client training in each round.}
\subsection{Data and Models}
\begin{figure*}
    \centering
    \includegraphics[scale=0.45]{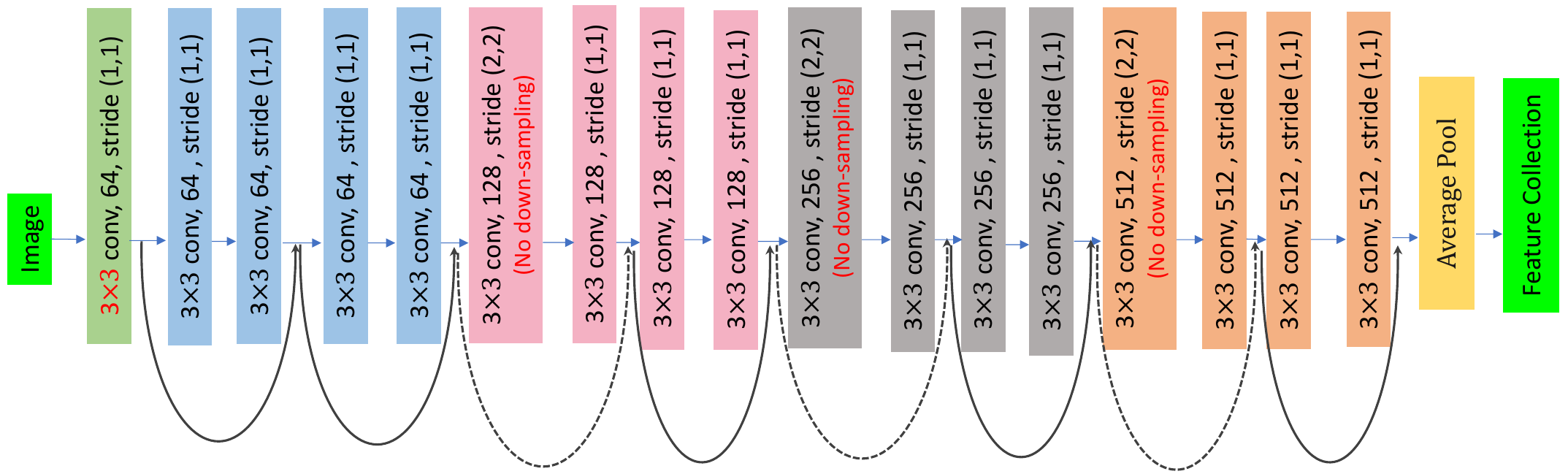}
    \caption{An overview of the modified ResNet-18 architecture. The first convolutional layer was adjusted to use a $3\times3$ kernel in place of the natural $7\times7$. The residual blocks~\cite{He2015ResNet} do not include the down-sampling operation (red) to maintain feature map dimensions.}    
    \label{fig:model_architecture}
\end{figure*}
\authone{To evaluate the effectiveness of the proposed methods, we conducted experiments on the CIFAR-10 dataset~\cite{krizhevsky2009learning}, a widely adopted benchmark in adversarial ML research. We preprocessed the CIFAR-10 training dataset using a series of transformation to enhance the diversity of the training data. First, the images are randomly cropped to a size of 32 x 32 pixels with a padding of 4 pixels. Next, a random horizontal flip is applied to each image, further preprocessing the dataset by simulating different orientations.} In our experiments, we employed the ResNet-18 architecture~\cite{He2015ResNet}. \authone{The ResNet-18 is a convolutional neural network designed for image classification tasks, consisting of 18 residual layers. The network begins with a convolutional layer with a kernel size of 3x3, a stride of 1, and padding of 1. The model is composed of four main blocks, each containing multiple residual layers. These residual layers include shortcut connections that mitigate the vanishing gradient problem in deep networks. As depicted in Figure~\ref{fig:model_architecture}, we applied two specific modifications to the original architecture.}\authone{While down-sampling can be applied at the beginning of block two, three, and four to reduce spatial dimensions, most residual layers within these blocks do not perform down-sampling. This design ensures that feature extraction is computationally efficient while maintaining spatial resolution. Additionally, ResNet-18 allows for flexibility in choosing activation functions. This enables the use of various activation functions.
In all experiments, the modified ResNet-18 architecture was trained from scratch.}
\subsection{Centralized AT} \label{sec:centralized_AT}
\authone{We followed the experimental setup outlined in Lin et al.~\cite{Li2020} to assess the ML model robustness of our centralized model on the CIFAR-10 dataset. We trained the model with PGD-based adversarial examples to enhance the ML model robustness against FGSM, C\&W, DeepFool, and PGD attacks. First, we initialized the adversarial samples by adding a random noise within the allowed perturbation range $(-\epsilon,\epsilon)$, where $\epsilon \approx 0.031$. Then, for a specified number of iterations $n_{iter} = 7$, we iteratively computed the gradient of the loss with respect to the inputs, took the sign of the gradient, and updated the adversarial examples by adding a scaled version of the gradient with the step size $\alpha \approx 0.00784$. To augment the training dataset with Gaussian noise examples, we generated noisy examples by adding Gaussian noise to clean examples. For each clean example, we defined a Gaussian distribution with a mean, $\mu_{GaussinNoise} = 0$, and a standard deviation, $\sigma_{GaussinNoise} = 0.1$. We sampled from the distribution and added the random noise to the clean sample. Training the model with Gaussian noisy examples enhances the ML robustness against random noise~\cite{lin2020secure}. We further modified the labels for the PGD adversarial and Gaussian based training samples using the soft label technique. The modified labels for the correct class is 0.955, indicating the confidence assigns to the correct lass. The incorrect classes are all other classes except the correct one. The soft labels among all incorrect classes are equal to  $\frac{\alpha_{SL}}{10}=0.005$.} 
The process for generating adversarial images was outlined in the Methodology section (Section~\ref{sec:method}). The centralized model was adversarially trained from scratch using the SGD optimizer with a learning rate of 0.001, momentum rate of 0.9, and weight decay rate of 0.0002 over 200 epochs. \authone{We employed categorical cross-entropy as the loss function because it is well-suited for multi-class classification tasks such as CIFAR-10. At test time, we measured the ML model robustness using FGSM, C\&W, DeepFool, and PGD attacks. The DeepFool iteratively computes minimal perturbations to move an input sample across the decision boundary of the trained model with an overshooting parameter $\epsilon_{DF} = 10^{-6}$. The iterative process continues until the model prediction on the perturbed input changes or the maximum number of iterations denoted as $\text{Max iterations}_{DF}$ is reached. We further tested the trained model with C\&W adversarial examples by setting the $\kappa_{CW}$ parameter of the C\&W attack objective function to 0, the learning rate $\alpha_{C\&W} = 0.01$, the number of iterations $\text{Max iterations}_{C\&W} = 10$, and the range for the regularization parameter $c \in [10^{-5}, 20.0]$. Prior to testing, Gaussian noise with a mean of 0 and a standard deviation of 0.1 was added to the test images. The parameters used for centralized AT training and testing phases are summarized in Table~\ref{tab:centralized_AT_params}. A comparison of our centralized AT method with Lin et al.'s approach~\cite{LinAT22} on the robust accuracy is presented in Table~\ref{table:centralized_AT_robustness}.} 

\begin{table}[h!]
    \centering
    \small\addtolength{\tabcolsep}{-1.0pt}
    \footnotesize
    \label{tab:param_summary}
    \begin{tabular}{| l | l |}
    \hline
    {\textbf{Parameters}} & {\textbf{CIFAR-10}}  \\ \hline
    $\mu_{GaussinNoise}$   & 0.00  \\ \hline
    $\sigma_{GaussinNoise}$                      & 0.1 \\ \hline
    $\alpha_{PGD}$              & 0.00784 \\ \hline
    $\alpha_{CW}$              & 0.01  \\ \hline 
    $\alpha_{SL}$               & 0.05  \\ \hline
    $\epsilon_{FGSM}$              & 0.031 \\ \hline
    $\epsilon_{PGD}$              & 0.031 \\ \hline
    $\epsilon_{DF}$              & $10^{-6}$ \\ \hline
    $\kappa_{CW}$                  & 0.0 \\ \hline 
    $\text{Max iterations}_{FGSM}$              & 1 \\ \hline
    $\text{Max iterations}_{PGD}$              & 7 \\ \hline
    $\text{Max iterations}_{DF}$              & 100 \\ \hline
    $\text{Max iterations}_{C\&W}$              & 10 \\ \hline
    The range of $\text{c}_{C\&W}$                  & $[10^{-5}, 20.0]$ \\ \hline
    Batch size              & 128 \\ \hline
\end{tabular}
\caption{Training and testing parameters}
\label{tab:centralized_AT_params}
\end{table}
\subsection{Federated AT with IID and Non-IID Data}
\authone{The following sections describe the preparation of both IID, non-IID data, and the local AT for our proposed federated AT method. All experiments were conducted using $K$ workers, where $K \in \{5, 10\}$. Adversarial examples were generated using the PGD algorithm, the Gaussian distribution, and soft labels using the same hyperparameters as in the centralized training setup (see Table~\ref{tab:centralized_AT_params}). The number of local training epochs was set to $E \in \{1, 3, 5\}$.}
\subsubsection{IID Data}
\authone{The IID sampling approach aims to split the CIFAR-10 training dataset into IID subsets for all FL clients. The training dataset were partitioned into equal subsets. Specifically, the number of data points in each each subset is equal to the total training's dataset size divided by the number of FL clients. There is no overlaps between all the subsets. All subsets have similar distributions of classes. For each subset, we randomly selected without replacement a fixed number of data points across the ten classes of the CIFAR-10 dataset, ensuring all the classes have a similar number of data points and each subset receives random examples. We assigned one subset per client. The IID sampling approach ensures that each client receives an equal-sized, class-balanced, randomly sampled subset of the dataset. Consequently, each client's local dataset is IID. Each client then trained its local model on its IID data. In Section~\ref{FAT_IID}, we compared the performance of our federated AT with the IID dataset to the centralized AT on the natural and robust accuracy.}
\subsubsection{Non-IID Data} \label{Non-IID_dataset}
\begin{figure*}
    \centering
    \includegraphics[width=1.0\linewidth]{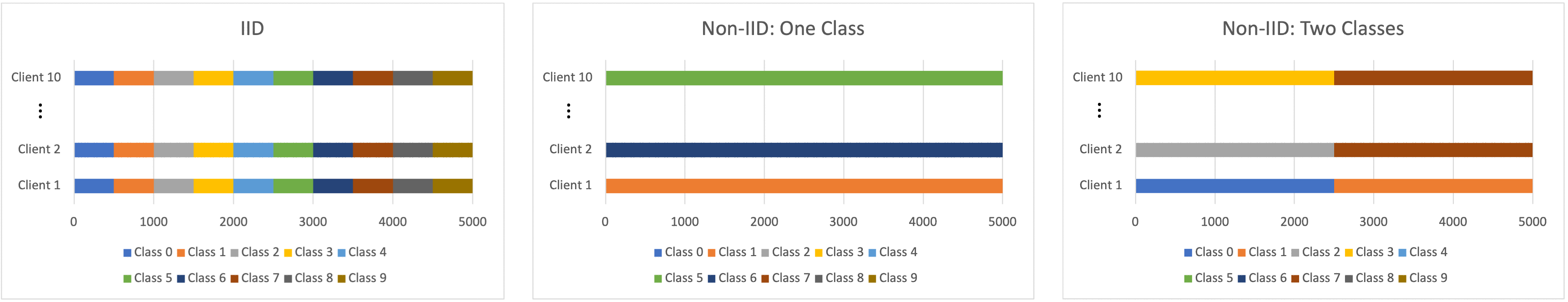}
    \caption{The IID and non-IID data split approaches among ten clients in the FL environment are shown. The dataset used is CIFAR-10, consisting of images from ten classes.}
    \label{fig:non_iid}
\end{figure*}

\authone{ Unlike the IID data split, where the training set is randomly divided into disjoint subsets with each subset following a uniform data distribution, the non-IID data split partitions the training dataset so that each partition exhibits a non-IID data distribution. This approach simulates real-world scenarios. In these scenarios, FL clients have access to data with skewed class distributions.} 
In our experiments, each client was randomly assigned one or two classes of data, as described in~\cite{ZhaoFLNoniidData}. \authone{In the one class non-IID strategy, each client is assigned data from only one class by dividing the dataset into partitions, where each partition contains data from a single class. We then randomly assigned one partition to each client. In the two-class strategy, each client is assigned data from two classes by randomly selecting two partitions per client.} The non-IID data sampling approach makes all clients' data distribution different from each other and skewed towards the assigned subset of classes. 
\authtwo{Figure~\ref{fig:non_iid} compares the IID and non-IID data split approaches across ten clients in the FL environment.}
\authone{Each client received their non-IID local datasets in addition to a portion of the globally shared dataset. Each client then performed the local AT on the combined training set. The effects of non-IID data on our federated AT approach, along with the data sharing strategy used to reduce the skewed class distribution across all clients, are discussed in Section~\ref{FAT_Non_IID}.}

\begin{table*}[ht!]
    \begin{tabular}{|p{3.7cm}| p{2.3cm}| p{1.6cm}|p{2.4cm} |p{2.0cm}| p{1.9cm}|}
    \hline
    Training Set & Model & Learning Rate &Test Set & \multirow{2}{1.0cm}{\textbf{Natural\\Accuracy}} & \multirow{2}{1.0cm}{\textbf{Robust\\Accuracy}} \\
    & & & & & \\
    & & & & & \\
    \hline 
    Natural examples & Customized ResNet-18 & Varied & FGSM & 99.26 & 3.37 \\
    \hline
    PGD examples & Customized ResNet-18 & Varied & FGSM & 78.17 & 47.05 \\
    \hline
    FGSM examples & Customized ResNet-18 & Varied & FGSM & 78.59 & 27.27 \\
    \hline
    PGD examples + Gaussian & Customized ResNet-18 & \authone{Fixed} & FGSM + Gaussian & 78.65 & 65.41 \\
    \hline
    PGD examples + Gaussian & Official ResNet-18 & Varied & FGSM + Gaussian & 70.17 & 44.82 \\
    \hline
    PGD examples + Gaussian & Customized ResNet-18 & \authone{Varied} & FGSM + Gaussian & 79.87 & 55.95 \\
\hline
\end{tabular}
\caption{Centralized AT - Natural and robust accuracy (\%) on the CIFAR-10 test set. Natural accuracy is the test accuracy of the trained ResNet-18 model on the clean test examples. Robust accuracy is the model's test accuracy on adversarial test examples generated by FGSM with $\epsilon = {8}/{255}$. }
\label{table:centralized_AT}
\end{table*}

\subsubsection{Local AT} \label{Local_training}
\authone{We adapted our centralized AT method to each client's local training process to ensure that trained local models are robust against adversarial attacks. The local training process involves each client using its own dataset, which includes both local data and half of the images from each class in the globally shared dataset. The training process iterates over a specified number of local epochs $E$, and for each epoch, the client's data is loaded in batches. During training, adversarial examples are generated using the Gaussian noise and PGD algorithm to augment the local training datasets. The labels are smoothed using the soft-labelling technique. The loss between the model's predictions and the soft labels is computed using the cross-entropy function. The local model's weights are updated using SGD. After training, the updated local model weights are returned to the server for aggregation. This process ensures that each client contributes to the global model. Our experiments are based on the assumption that every client participates in each round of communication, following the configuration outlined in Shah et al.~\cite{shah2021adversarial}.}
\subsection{Results}
\authone{This section summarizes the empirical results of the trained models when tested on natural and adversarial images using our proposed centralized and federated AT algorithms. Our methods achieved notable improvements on natural and robust accuracy in the centralized and federated environments, with both IID and non-IID data scenarios. The evaluation highlights the model’s generalization capacity on natural examples and the ML model robustness against adversarial examples at test time.} 
\subsubsection{Centralized AT} \label{Centralized_training_result}
\authtwo{We first trained the customized ResNet-18 model on CIFAR-10 using standard training and achieved natural and robust accuracy of 99.26\% and 3.37\%, respectively. Next, we conducted centralized AT using PGD examples, achieving the natural accuracy of 78.17\% and robust accuracy of 47.05\%. On the other hand, when PGD examples are replaced by FGSM examples with $\epsilon = {8}/{255}$, the robust accuracy decreased from 47.05\% to 27.27\%. This demonstrates that PGD examples, produced by iterative local search process around the neighborhood of the natural examples, are more effective. We further improved the robust accuracy to 65.41\% by adding examples with Gaussian noise to the training dataset. When the same AT approach was implemented on  the official ResNet-18~\cite{He2015ResNet}, performance decreased in both natural accuracy (70.17\% vs 78.65\%) and robust accuracy (44.82\% vs 65.41\%). These results are presented in Table \ref{table:centralized_AT}. Using a varied learning rate, as proposed in Section~\ref{sec.lr}, instead of a fixed learning rate, slightly increased the natural accuracy (from 78.17\% to 79.87\%) but dramatically decreased the robust accuracy (from 65.41\% to 55.95\%). Therefore, we adopted the centralized AT approach with the customized ResNet-18 model trained with a fixed learning rate for comparison with Lin, et al.~\cite{LinAT22}. The results are shown in Table \ref{table:centralized_AT_robustness}. The robust accuracy reported in Table \ref{table:centralized_AT_robustness} demonstrates the improvement of our proposed AT approach over Lin, et al.`s method~\cite{LinAT22} against different white-box attacks. Our method achieved higher robust accuracy against the FGSM, C\&W, and DeepFool attacks highlighting the effectiveness of our centralized AT method.}
\begin{table}[H]
    \centering
    \begin{tabular}{|l|l|l|l|}
    \hline
        Training & FGSM & C\&W & DeepFool \\ \hline
        Lin, et al.~\cite{LinAT22} & 47 & 78 & 36  \\ \hline
        Our method & 65.41 & 81 & 83 \\ \hline
    \end{tabular}
    \caption{Centralized AT - Robust accuracy (\%) on the CIFAR-10 test dataset under various white-box attacks.}
    \label{table:centralized_AT_robustness}
\end{table}

\subsubsection{Centralized AT with Activation Functions} \label{variants_of_ReLU__AF}
\authone{In this section, we evaluated how the set of eleven activation functions combined with our centralized AT process impacts the ML model robustness. We conducted two experiments using the modified ResNet-18 model. The first experiment extended our experiment in Section~\ref{Centralized_training_result} with the additional activation functions. In the second experiment, instead of using a fixed learning rate, we applied a linearly decaying learning rate as outlined in Section~\ref{sec:method}. Both experiments used the same setup as in Section~\ref{Centralized_training_result}, where models were trained for 200 epochs with an initial learning rate of 0.001. To measure \authtwo{the} ML model robustness \authtwo{of the resulting models}, we only use adversarial examples generated by the FGSM method with the maximum perturbation $\epsilon = 8/255$, as ReLU model’s robust accuracy is already significantly low at 3.37\% without the centralized AT. The results of these two experiments are summarized in Table~\ref{table:AF_r200_e1_fixed_lr}.
\begin{itemize}
  \item \textbf{Fixed learning rate results:} With a fixed learning rate of 0.001, ReLU achieved the highest robust accuracy (67.96\%). RReLU attained a similar robust accuracy of 67.00\% but slightly outperformed ReLU in natural accuracy (78.71\% vs. 77.08\%). SELU obtained only 40.08\% robust accuracy. Some good values for the $\alpha$ parameter that could result in improved robust accuracy are -0.5, 0 and 0.5 as suggested in~\cite{dai2022parameterizing}. We observed that SiLU, which is a smooth approximation of ReLU, performed worse than ReLU by achieving only 38.85\% robust accuracy. However, as shown in ~\cite{dai2022parameterizing}, SiLU outperformed ReLU in both robust and natural accuracy by introducing a parameter $\alpha = 56.23$. The remaining activation functions - TELU, GeLU, Mish, HardSiLU, CeLU, and HardTanh performed worse than the ReLU in terms of the robust accuracy, ranging from 38.25\% achieved by GeLU to 52.40\% obtained by TELU. While TELU's robust accuracy was lower than ReLU, it improved natural accuracy from 77.08\% achieved by ReLU to 80.28\%. This observation contrasts with findings in~\cite{fernandez2024stable} in terms of natural accuracy. As shown in~\cite{fernandez2024stable}, ReLU performed better than TELU when using SGD with a decayed learning rate.
  \item \textbf{Varying learning rate results:} Under the varying learning rate scheme, ReLU obtained a robust accuracy of 50.67\% while RReLU, TELU, Softplus, and CeLU (learnable parametric activation function) showed a significant improvement in robust accuracy compared to the ReLU baseline. In Table~\ref{table:AF_r200_e1_fixed_lr}, RReLU, TELU, Softplus, and CeLU obtained 57.60\% (+6.93\%), 53.74\% (+3.07\%), 58.86\% (+8.19\%), and 57.64\% (+6.97\%) robust accuracy. The '+' values indicate the percentage increase in robust accuracy compared to the ReLU baseline of 50.67\%. However, RReLU, Softplus, and CeLU enhanced the robust accuracy at the cost of lowering natural accuracy. Only TELU—a complex activation function combining identity, tanh, and exponential functions—did not exhibit a significant trade-off between natural and robust accuracy. TELU improved robust accuracy from 50.67\% achieved by ReLU to 53.74\%, with only a marginal drop in natural accuracy (80.90\% obtained by ReLU to 80.68\%). This result aligned with findings that ReLU surpassed TELU in natural accuracy when trained with SGD with a decayed learning rate~\cite{fernandez2024stable}. GeLU, Mish, SiLU, HardSiLU, and HardTanh had worse robust accuracy than ReLU, ranging from 37.54\% achieved by GeLU to 47.55\% obtained by HardTanh. 
  \item In summary, ReLU remained the best choice for our centralized AT process with a fixed learning rate. For varying learning rates, TELU was a strong alternative, because it achieved a good trade-off between natural and robust accuracy. However, our adversarially trained models using Softplus, CeLU and RReLU suffered from a worse trade-off between natural accuracy and robust accuracy. In future work, we will focus on identifying a more optimal trade-off between natural accuracy and robust accuracy as demonstrated in~\cite{balaji2019instance}. \authtwo{Note that RReLU's performance in both the fixed and varying learning rate is similar to ReLU.} We further explored the impacts of activation function on the ML model robustness through our centralized AT in Section~\ref{Discussion}. Specifically, we set PGD's number of iterations equal to 7, 14, and 21.
\end{itemize}}

\begin{table*}[!ht]
    \centering
    \renewcommand{\arraystretch}{1}
    \begin{tabular}{|p{2.8cm} | p{2cm} | p{2.0cm} | p{1.7cm} | p{1.7cm} | p{1.7cm} | p{1.7cm}|}
        \hline
            \multicolumn{3}{|c|}{} & \multicolumn{2}{c|}{Fixed learning rate} & \multicolumn{2}{c|}{Varying learning rate}\\
        \hline
            Model  & Dataset & Activation Function & Natural Accuracy (\%) & Robust Accuracy (\%) & Natural Accuracy (\%) & Robust Accuracy (\%)\\ 
        \hline
            \multirow{11}{5em}{The customized ResNet-18} & \multirow{11}{5em}{CIFAR-10} & ReLU & 77.08 & \textbf{67.96}
            & 80.90 & 50.67 \\ 
            \cline{3-7}
            & & RReLU & \textbf{78.71}  & 67.00  & 78.04 & \textbf{57.60} \\ 
            & & TELU & \textbf{80.28}  & 52.40 & 80.68 & \textbf{53.74} \\ 
            & & GeLU & \textbf{80.27}  & 38.25 & \textbf{81.86} & 37.54 \\ 
            & & Mish & \textbf{82.32}  & 42.80 & \textbf{82.01} & 39.63 \\ 
            & & SiLU & \textbf{81.72}  & 38.85 & 80.67 & 45.45 \\ 
            & & Softplus & \textbf{79.39}  & 49.72 & 74.92 & \textbf{58.86} \\ 
            & & CeLU & \textbf{82.61}  & 42.02 & 77.85 & \textbf{57.64} \\ 
            & & HardSiLU & \textbf{78.36}  & 42.84 & \textbf{81.43} & 43.62 \\ 
            & & SELU & \textbf{78.42}  & 40.08 & 76.22 & 44.14 \\ 
            & & HardTanh & 72.56  & 58.54 & 69.22 & 47.55 \\ 
        \hline
    \end{tabular}
    \caption{Centralized AT - natural and robust accuracy for various activation functions on the CIFAR-10 test dataset where the number of epochs is 200. \authtwo{The values \authone{in bold }\authone{are those accuracies that are} higher than the corresponding accuracies\authone{, obtained by using} ReLU. We can observe that no other activation function performs better than ReLU, in terms of both natural and robust accuracy.}}
    \label{table:AF_r200_e1_fixed_lr}
\end{table*}

\subsubsection{Federated AT with IID Data} \label{FAT_IID}
\authone{In this section, we focus on evaluating the performance of Federated AT in an ideal setting where the local data across all clients is IID. The goal was to achieve robust accuracy comparable to the centralized AT case. The results, as shown in Table~\ref{table:FAT_AT-IID}, indicate that federated AT with IID data can indeed achieve the ML model robustness against C\&W, DeepFool, and PGD that is comparable to the centralized setting. For example, with five clients, the robust accuracy for C\&W was similar to the centralized model. This aligns with observations from other studies including Zizzo et al.~\cite{zizzo2020fat} and Shah et al.~\cite{shah2021adversarial}, who also reported that federated AT with IID data yields performance close to the centralized scenario. When the number of clients were ten, we observed a similar trend in robust accuracy across C\&W, DeepFool, and PGD with the five client case. Specifically, the federated approach achieved the robust accuracy within 5\% and 7\% of the centralized AT's robust accuracy for the C\&W and DeepFool attacks, respectively. For the PGD attack, the federated approach with ten clients outperformed the centralized case by 5\%. These findings suggest that in the ideal IID data scenario, the decentralized nature of FL does not inherently impede the training of robust models against adversarial examples.} 
\begin{table}[!ht]
    \centering
    \begin{tabular}{| l | l | l | l | l | l | l |}
    \hline
        \# Clients & Natural & FGSM & C\&W & DF & PGD\\ \hline
        K = 5 & 80.76  & 63.07  & 81 & 79 & 71 \\ \hline
        K = 10 & 66.23  & 51.51  & 76 & 76 & 77 \\ \hline
    \end{tabular}
    \caption{IID Federated AT - Robust accuracy (\%) on the CIFAR-10 test dataset under various white-box attacks. (Note: $R = 100$ and $E = 3$. DF refers to DeepFool.)}
    \label{table:FAT_AT-IID}
\end{table}
\subsubsection{Federated AT with Non-IID Data}
\label{FAT_Non_IID} 
\authone{In this section, we examine the challenges of applying federated AT when the data across all participating clients is non-IID and the effectiveness of the data sharing technique \cite{ZhaoFLNoniidData} in mitigating the adverse effect of non-IID data on federated AT. The experimental results are presented in Table~\ref{table:oneclass_non_iid_ed_AT_robustness_r100} and Table~\ref{table:twoclass_non_iid_ed_AT_robustness_r100}. In these tables, we compared the natural and robust accuracy of our federated AT method on IID and non-IID data with and without using the data sharing method for the local training~\cite{ZhaoFLNoniidData}. In contrast to the IID data setting presented in Section~\ref{FAT_IID}, the one-class and two-class non-IID data present significant decrease in performance for the federated AT. For instance, when comparing the IID case to the two-class non-IID case within ten clients, the natural accuracy dropped significantly from 66.23\% (Table~\ref{table:FAT_AT-IID}) to 57.82\% (Table~\ref{table:twoclass_non_iid_ed_AT_robustness_r100}). This data heterogeneity also severely impacts the robust accuracy against adversarial attacks. We observed a large performance gap in the robust accuracy when comparing the one-class non-IID case with ten clients (Table~\ref{table:oneclass_non_iid_ed_AT_robustness_r100}) to the IID case (Table~\ref{table:FAT_AT-IID}). There was a decrease of 23\% (from 76\% to 53\%) and 31\% (from 77\% to 48\%) in the robust accuracy against C\&W and PGD attacks, respectively. This degradation in both the natural accuracy and the robust accuracy of the global model can be attributed to the weight divergence phenomenon introduced by Zhao et al.~\cite{ZhaoFLNoniidData}, which can be measured by comparing the Earth mover's distance difference in local model weights resulting from training on skewed local data distributions.}
\authone{To mitigate the adverse effects of non-IID data on the federated AT, we adopted a data-sharing strategy proposed by Zhao et al.~\cite{ZhaoFLNoniidData}. This approach involves creating a globally shared dataset that contains some examples from each class and distributing the shared dataset to each client. Specifically, in our experiments, we set $\alpha$ = 0.5 as suggested in~\cite{ZhaoFLNoniidData} for the optimal natural accuracy. In particular, we randomly sampled 500 images per class in the global shared dataset. After creating the global training subset, we divided the unselected training examples into multiple non-IID partitions. Each non-IID partition was assigned to each client for local AT. Each client received its non-IID local data partition and, in addition, a random sample of 500 images per class from the globally shared dataset. The details of generating the global shared dataset and crafting the non-IID data for each client had been described in Section~\ref{sec:data_sharing}.}\authone{ Table~\ref{table:oneclass_non_iid_ed_AT_robustness_r100} and Table~\ref{table:twoclass_non_iid_ed_AT_robustness_r100} demonstrate that using this technique can significantly improve both natural and robust accuracy for one-class and two-class non-IID data. 
In the case in which each client in federated AT only has data from one-class, the global model trained without using the global shared training subset performed poorly in both natural accuracy and robust accuracy. Specifically, the model's natural accuracy was only 10.97\%, and its robust accuracy against FGSM, C\&W, DeepFool, and PGD was also low, at 1.61\%, 11\%, 12\%, and 11\% ,respectively. These results were detailed in Table~\ref{table:oneclass_non_iid_ed_AT_robustness_r100}. On the other hand, sharing a small fraction of data allowed for an increase in both natural and robust accuracy. Specifically, the natural accuracy increases from 10.97\% to 67.42\% and the robust accuracy against C\&W attack achieves a significant improvement from 11\% to 54\%, as shown in Table~\ref{table:oneclass_non_iid_ed_AT_robustness_r100}.
In the two-class non-IID scenario, similar to the one-class non-IID case, using the data sharing technique led to better performance in terms of both natural and robust accuracy. The specific improvements for these two-class setups are detailed in Table~\ref{table:twoclass_non_iid_ed_AT_robustness_r100}.}
The result demonstrates that even sharing a small portion of the globally shared dataset can effectively address the challenge posed by non-IID data in achieving robust FL global models against the adversarial attacks.
\begin{table*}[!ht]
    \centering
    \begin{tabular}{|p{6cm}|p{1.5cm}|p{1.5cm}|p{1.5cm}|p{1.5cm}|p{1.5cm}|p{1.3cm}|}
    \hline
        Training & Natural & FGSM & C\&W & DeepFool & PGD\\ \hline
        One class non-IID federated AT \textbf{without} data sharing & 10.97\%  & 1.61\%  & 11\% & 12\% & 11\% \\ \hline
        One class non-IID federated AT \textbf{with} data sharing  & 67.42\%  & 41.18\%  & 53\% & 47\% & 48\% \\ \hline
    \end{tabular}
    \caption{One class non-IID Federated AT - Robust accuracy (\%) on the CIFAR-10 test dataset under various white-box attacks, where K = 10, R = 100, and E = 1.}
    \label{table:oneclass_non_iid_ed_AT_robustness_r100}
\end{table*}

\begin{table*}[!ht]
    \centering
    \begin{tabular}{|p{6cm}|p{1.5cm}|p{1.5cm}|p{1.5cm}|p{1.5cm}|p{1.5cm}|p{1.5cm}|}
    \hline
        Training & Natural & FGSM & C\&W & DeepFool & PGD\\ \hline
        Two-class non-IID federated AT \textbf{without} data sharing & 
        57.82\%  & 54\%  & 57\% & 62\% & 59\% \\ \hline
        Two-class non-IID federated AT \textbf{with} data sharing  & 85.04\%  & 63.97\%  & 72\% & 71\% & 67\% \\ \hline
    \end{tabular}
    \caption{Two-class non IID Federated AT - Robust accuracy (\%) on the CIFAR-10 test dataset under various white-box attacks, where K = 10, R = 100, and E = 3.}
    \label{table:twoclass_non_iid_ed_AT_robustness_r100}
\end{table*}

\begin{table*}[!ht]
    \centering
    \small
    \renewcommand{\arraystretch}{1}
    \begin{tabular}{|p{1.5cm} | p{1.3cm} | p{1.5cm} | p{1.4cm} | p{1.4cm} | p{1.4cm} | p{1.4cm} | p{1.4cm} | p{1.4cm}|}
        \hline
            \multicolumn{3}{|c|}{} & \multicolumn{2}{c|}{PGD-7} & \multicolumn{2}{c|}{PGD-14} & \multicolumn{2}{c|}{PGD-21}\\
        \hline
            Model & Dataset & Activation Function & Natural Accuracy & Robust Accuracy & Natural Accuracy & Robust Accuracy & Natural Accuracy & Robust Accuracy\\ 
        \hline
            \multirow{11}{5em}{The customized ResNet-18} & \multirow{11}{5em}{CIFAR-10} & ReLU & 79.37 & 49.88 & 76.69 & 54.94 & 74.37 & 54.36 \\
            \cline{3-9}
            & & RReLU & 75.85 & 49.28 & 74.56  & \textbf{55.67}  & \textbf{76.38}  & 39.28  \\
            & & TELU  & 77.09 & \textbf{50.29} & \textbf{78.98}  & 41.08  & \textbf{75.60}  & 47.72  \\
            & & GeLU  & \textbf{80.07} & 42.96 & \textbf{78.93}  & 46.73  & \textbf{78.11}  & 46.97  \\
            & & Mish  & \textbf{80.93} & 38.07 & \textbf{76.94}  & 42.44  & \textbf{78.72}  & 43.07  \\
            & & SiLU  & \textbf{79.72} & 46.49 & 74.47  & 40.77  & \textbf{75.04}  & 43.95  \\
            & & Softplus & 71.68 & \textbf{55.71} & 75.49  & 49.19  & 71.33  & \textbf{55.92}  \\
            & & CeLU & 73.82 & \textbf{56.70} & 74.19  & \textbf{63.86}  & \textbf{76.87}  & \textbf{61.35 } \\
            & & HardSiLU & 78.51 & 44.94 & \textbf{78.08}  & 47.19  & \textbf{76.19}  & 45.28  \\
            & & SELU  & 69.24 & 29.89 & 72.34  & \textbf{56.19}  & 71.47  & \textbf{60.05}  \\
            & & HardTanh & 59.18 & \textbf{59.09} & 60.33  & \textbf{55.02}  & 61.09  & \textbf{55.15}  \\
          \hline
    \end{tabular}
    \caption{Centralized AT - Natural and Robust accuracy (\%) for various activation functions on the CIFAR-10 test dataset where the number of epochs is 100 and number of PGD updates are 7, 14 and 21. \authtwo{The values in bold are those accuracies that are higher than the corresponding accuracies, and they are obtained by using ReLU. ReLU still stand as the best option for maintaining both natural and robust accuracy, expect for PGD-21 case where CELU performs slightly better than ReLU.}}
    \label{tab:PGD_iterations}
\end{table*}
\subsubsection{Discussion}
\label{Discussion}
\begin{figure}
    \centering
    \includegraphics[width=\linewidth]{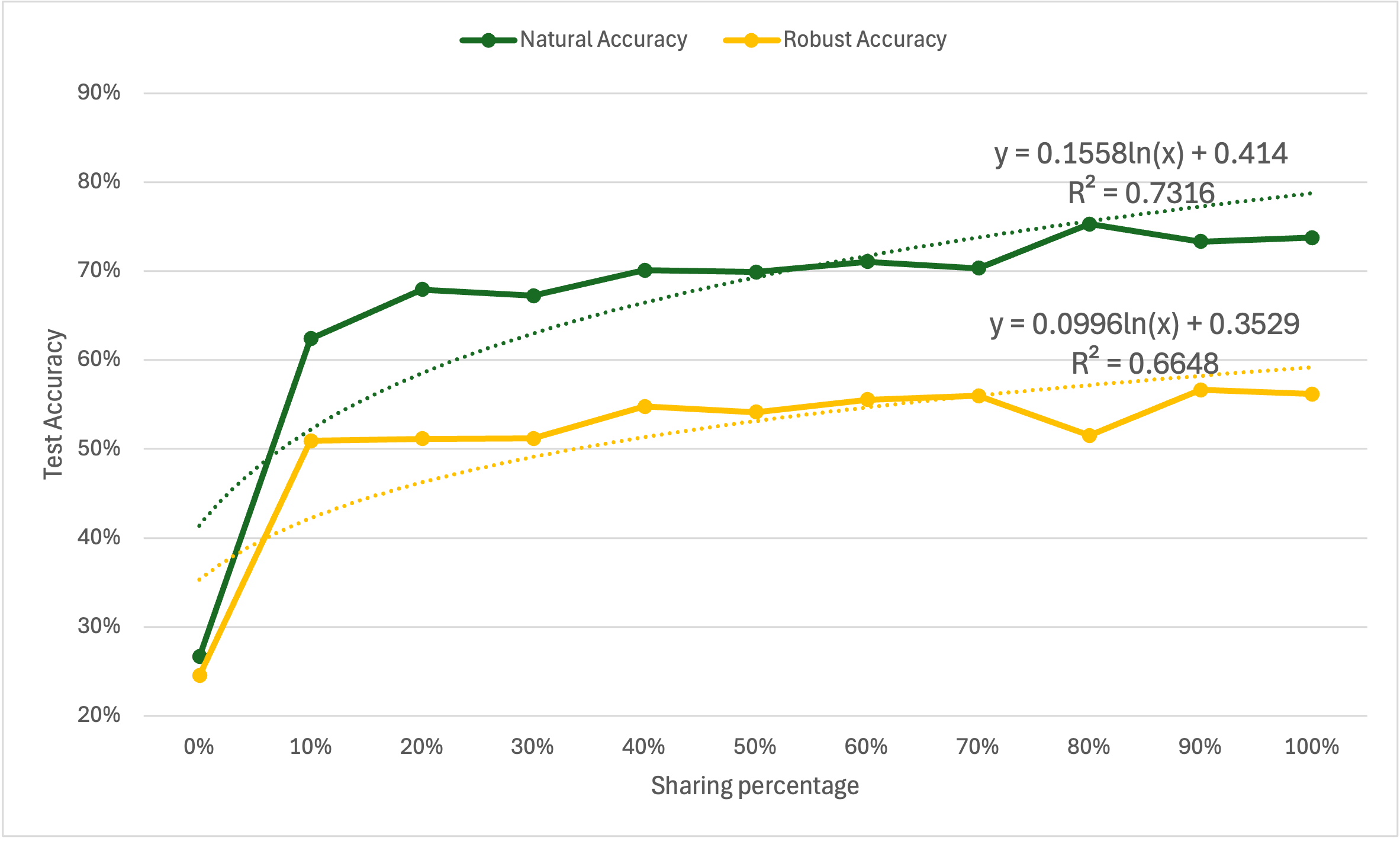}
    \caption{The plot illustrates how the natural and robust accuracy in the FL setting with non-IID data varies based on the sharing percentage of IID data. We fit prediction models for both natural and robust accuracy with respect to the sharing percentage.}
    \label{fig:sharing}
\end{figure}

\authone{In this section, we further explored our centralized and federated AT. For the centralized AT method, we study the impacts of the number of iterations used in the PGD algorithm on robust accuracy. Additionally, we examined the effects of our federated AT method, focusing on varying data sharing ratios.}
\begin{itemize} 
    \item \authone{ \textbf{Training with stronger adversarial examples}: Given the results from Section~\ref{variants_of_ReLU__AF}, we hypothesized that our centralized AT with activation functions required stronger adversarial examples. Specifically, we generated stronger adversarial examples by doubling and tripling the number of PGD iterations while fixing the maximum perturbation and the PGD's step size~\cite{dong2018boosting, MadryMSTV17}. We adversarially trained the modified ResNet-18 with various activation functions for 100 epochs, instead of the 200 epochs. We set the maximum perturbation to $\frac{8}{255}$ and the PGD's step size to $\frac{2}{255}$ while we experimented with three different iteration counts\authtwo{:}7, 14, and 21\authtwo{, which are} referred to as PGD-7, PGD-14, and PGD-21, respectively. By increasing the number of iterations, we can thoroughly explored the adversarial input landscape. We used FGSM to evaluate the ML model robustness. As shown in Table~\ref{tab:PGD_iterations}, PGD-14 with RReLU, CeLU, SELU, and HardTanh outperformed PGD-14 with ReLU in terms of robust accuracy. We emphasized the robustness improvement in the Table~\ref{tab:PGD_iterations} by highlighting the corresponding figures in bold. Notably, CeLU and HardTanh with PGD-7, PGD-14, and PGD-21 had better robust accuracy than ReLU-7, ReLU-14, and ReLU-21, respectively. Interestingly, as the number of iteration increased, the robust accuracy improvement was not stable. For example, CeLU-21's robust accuracy was better than ReLU-21 but less than CeLU-14. In comparison to the ReLU baseline for PGD-14 and PGD-21, SELU consistently improved the ML model robustness. For instance, with PGD-21, SELU increased the ML model robustness from 54.36\% to 60.05\%. When training with stronger adversarial examples, RReLU with PGD-14 outperformed ReLU-14 but performed worse than ReLU-21 when training with PGD-21 adversarial examples. Both RReLU, SELU, and CeLU exhibited lower natural accuracy compared to their ReLU counterparts, except for CeLU with PGD-21. For example, SELU and CeLU with PGD-14 achieved a natural accuracy 72.34\% and 74.19\% compared to 76.69\% obtained by the ReLU. We did not observe any improvement in the ML model robustness with GELU, Mish, SiLU, and HardSiLU. In summary, when the number of PGD's iterations were 14 and 21, RReLU, Softplus, CeLU, SELU, and HardTanh enhanced the ML model robustness; however, they failed to maintain the natural accuracy of ReLU. Additionally, during the training phase, we noted that the final training accuracies for ReLU, SELU, and CeLU  were significantly higher than their natural test accuracy, indicating the presence of overfitting, which can be mitigated by early stopping~\cite{rice2020overfitting}}.
    \item \authone{\textbf{Various percentage sharing:} As we demonstrated in Section~\ref{FAT_Non_IID}, the data-sharing method mitigated the reduction in robust accuracy for FL with non-IID datasets. There remains the problem of choosing the optimal sharing percentage for the highest robust accuracy. Intuitively, the robust accuracy increases when the size of the sharing dataset increases. We verified the hypothesis by conducting federated learning with the two-class non-IID data while varying the data sharing percentage. We trained the modified ResNet-18 model with ReLU on the two-class non-IID CIFAR-10 dataset with our federated AT method. We fixed K = 5, R = 50, E = 3, and experimented with the data sharing percentage starting from 0\%, 10\% up to 100\%. We measured the ML model robustness using the FGSM attack. Table~\ref{tab:data_share} shows the change in the natural and robust accuracy when the data sharing percentage increases. We found that increasing the data sharing percentage did not always increase the robust test accuracy. Specifically, we achieved the highest test robust accuracy of 56.64\% when the data sharing was 90\%. However, the robust accuracy decreased as the sharing percentage approached 50\%, 80\%, and 100\%.} \authtwo{Figure~\ref{fig:sharing} illustrates the natural and robust accuracy with respect to the sharing percentage. 
    \authone{Next, we modeled the robust and natural test accuracy changes as a function of the data sharing percentage}. Based on our experimental results, we fitted regression models to predict robust and natural accuracy, applying a natural log transformation to the sharing percentage to capture the pattern reflected in the data. The fitted regression model for robust accuracy is $ y_r = 0.0996 \cdot \ln(x) + 0.3529$, where $y_r$ represents robust accuracy and $x$ denotes the sharing percentage. This model gave an $R^2$ value of 0.6648, indicating a moderate fit to the data. We also provided a regression model to predict natural accuracy based on the sharing percentage: $ y_n = 0.1558 \cdot \ln(x) + 0.414$, where $y_n$ represents natural accuracy and $x$ the sharing percentage. This model has a better fit than the one for robust accuracy, with an $R^2$ value of 0.7316.}
    \item \authfour{\textbf{Non-IID data with label skewness:} Chen et al.~\cite{chen2022calfat} studied the challenge of non-IID data when there are few training examples for some target classes in each client's training set. Due to the non-IID data with label skewness, the global models trained by traditional Federated AT methods only learn patterns given by the majority target classes. Thus, their adversarially trained global models are not likely to generalize well on those minority target classes~\cite{chen2022calfat}. To overcome this challenge, Chen et al.~\cite{chen2022calfat} proposed a calibrated federated AT method known as CalFAT. CalFAT aims to directly balance the local logit vectors based on some class-wise probability information obtained from the training sets. One of the advantages of this approach is that it can be easily adapted to different datasets. The major difference between our training algorithm and Chen et al.~\cite{chen2022calfat} lies in how we address the challenge of non-IID data. Our federated AT algorithm tackles this issue through the data sharing mechanism~\cite{ZhaoFLNoniidData}. The global dataset obtained by the data sharing mechanism mitigates the non-IID data with label skewness challenge. Table~\ref{table:CalFAT_vs_OurFAT} shows that under the same experimental setting, our federated AT achieved notable advancement in both natural and robust accuracy over the CalFAT. Our federated AT achieved a robust accuracy of 50.17\%, significantly outperforming CalFAT's one by 43.3\%. Additionally, our adversarially trained model’s natural accuracy was better than CalFAT, with scores of 64.69\% compared to 68.52\%. We hypothesize that this enhancement comes from the global shared dataset. The global shared dataset enables all the local models to be trained on nearly IID training sets before the aggregation step. This, in turn, allows the aggregated model to eventually converge to a generalized and robust ML one. When we reduced the label skewness in the non-IID data by setting Dirichlet($\beta = 0.4$) and increased the number of local epochs from one to three, we observed notable improvements. Table~\ref{table:skewed_label_non_iid_ed_AT_robustness_K_5} demonstrated that our federated AT improved the natural accuracy from 68.52\% for Dirichlet($\beta = 0.1$) to 72.07\% and 81.27\% for Dirichlet($\beta = 0.4$) and $E = \{1, 3\}$, respectively. Additionally, the robust accuracy increased from 50.17\% to 56.24\% and 54.66\% compared to the Dirichlet($\beta = 0.1$) results. Furthermore, Table~\ref{table:skewed_label_non_iid_ed_AT_robustness_K_10} illustrated the stability of our federated AT method under the label skewness using the Dirichlet($\beta = 0.4$). This stability was observed when we increased the number of clients from five to ten. Specifically, our federated AT with ten clients achieved a worse performance compared to the case when the number of clients is five. We conjectured that this is because there are fewer training examples for each client, making it difficult to achieve better performance. However, our federated AT with both five and ten clients still enhanced both the natural accuracy and robust accuracy. This improvement is noted when compared to the worst case non-IID data with label skewness at Dirichlet($\beta = 0.1$) as shown in Table~\ref{table:CalFAT_vs_OurFAT}. The higher natural accuracy combined with the stronger ML model robustness suggests that the proposed federated AT technique is a useful tool. It is valuable to develop robust ML models based on non-IID data with label skewness.}
\end{itemize}
\begin{table}[H]
    \centering
    \begin{tabular}{| l | l | l |}
    \hline
        Training & Natural & FGSM\\ \hline
        CalFAT~\cite{chen2022calfat}  & 64.69\%  & 35.02\% \\ \hline
        Our federated AT  & 68.52\%  & 54.66\% \\ \hline
    \end{tabular}
    \caption{Non-IID Federated AT - Robust accuracy (\%) on the CIFAR-10 test dataset under various white-box attacks, where K = 5, R = 150, E = 1, Dirichlet ($\beta$= 0.1).}
    \label{table:CalFAT_vs_OurFAT}
\end{table}
\begin{table*}[!ht]
    \centering
    \begin{tabular}{|p{6cm}|p{1.5cm}|p{1.5cm}|p{1.5cm}|p{1.5cm}|p{1.5cm}|p{1.3cm}|}
    \hline
        Training & Natural & FGSM & C\&W & DeepFool & PGD\\ \hline
        Our federated AT with $E = 1$  & 72.07\%  & 56.24\%  & 77\% & 75\% & 70\% \\ \hline
        Our federated AT with $E = 3$  & 81.27\%  & 54.66\%  & 79\% & 74\% & 65\% \\ \hline
    \end{tabular}
    \caption{Non-IID Federated AT - Robust accuracy (\%) on the CIFAR-10 test dataset under various white-box attacks, where K = 5, R = 200, E = {1, 3}, Dirichlet ($\beta$ = 0.4).}
    \label{table:skewed_label_non_iid_ed_AT_robustness_K_5}
\end{table*}
\begin{table*}[!ht]
    \centering
    \begin{tabular}{|p{6cm}|p{1.5cm}|p{1.5cm}|p{1.5cm}|p{1.5cm}|p{1.5cm}|p{1.3cm}|}
    \hline
        Training & Natural & FGSM & C\&W & DeepFool & PGD\\ \hline
        Our federated AT with $E = 1$  & 73.59\%  & 51.42\%  & 78\% & 79\% & 78\% \\ \hline
        Our federated AT with $E = 3$  & 78.56\%  & 55.53\%  & 78\% & 79\% & 68\% \\ \hline
    \end{tabular}
    \caption{Non-IID Federated AT - Robust accuracy (\%) on the CIFAR-10 test dataset under various white-box attacks, where K = 10, R = 200, E = {1, 3}, Dirichlet ($\beta$ = 0.4).}
    \label{table:skewed_label_non_iid_ed_AT_robustness_K_10}
\end{table*}
\begin{table}[ht]
    \centering
   {  
    \renewcommand{\arraystretch}{1.2}
    \begin{tabular}{| p{1.6cm} | p{1.6cm} | p{1.6cm} |}
        \hline
        \multirow{2}{1.5cm}{The data sharing\\percentage} & \multirow{2}{1.0cm}{Natural\\Accuracy} & \multirow{2}{1.0cm}{Robust\\Accuracy} \\
           & & \\
           & & \\
          \hline 
          0  & 26.67 & 24.54 \\
           \hline 
          10  & 62.4 & 50.92 \\
           \hline 
          20  & 67.93 & 51.14 \\
           \hline 
          30  & 67.23 & 51.99 \\
           \hline 
          40  & 70.09 & 54.79 \\
           \hline 
          50  & 69.91 & 54.13 \\
           \hline 
          60  & 71.05 & 55.54 \\
           \hline 
          70  & 70.33 & 55.99 \\
           \hline 
          80  & 75.29 & 51.51 \\
           \hline 
          90  & 73.31 & 56.64 \\
           \hline 
          100  & 73.77 & 56.16 \\
         \hline
    \end{tabular}}
    \caption{ Two-class non-IID Federated AT - Natural and Robust Accuracy (\%) on the CIFAR-10 test dataset under different data sharing percentages (\%), where K = 5, R = 50, and E = 3.}
    \label{tab:data_share}
 \end{table}
\section{Conclusion and Future Work} 
\label{sec:conclusion}
\authtwo{While existing studies have explored various activation functions, the impact of these functions on the ML model robustness through AT in centralized environments remains unclear. Furthermore, the effects of IID and non-IID data on ML robustness through AT in decentralized environments are not well understood.\authone{In this paper,} we address\authone{ed} this research gap by proposing an advanced AT approach in both centralized and decentralized environments. \authone{Specifically}, we enhanced the AT approach introduced by Lin, et al.~\cite{LinAT22} for a centralized environment and extended it to FL in a decentralized environment. The proposed approach achieved significantly better robust accuracy compared to Lin, et al.~\cite{LinAT22}. Our extensive experiment\authone{s} on eleven activation functions shows that due to its smoothness, TELU was a strong alternative to ReLU for fixed learning rate. In the fixed learning rate setting, ReLU remained the best option for our centralized AT. In decentralized environments, we explore\authone{d} the ML model robustness with both IID and non-IID data. To improve the ML robustness with non-IID data, we \authone{incorporated a data sharing approach for non-IID data, where we} required all clients to share a portion of their unique data to mitigate the data disparities among all clients. \authfour{In terms of robust accuracy, the \authtwo{proposed} federated AT method, using the data sharing mechanism, considerably surpassed CalFAT~\cite{chen2022calfat}.} In general, our experimental results on CIFAR-10 demonstrated that the ML model robustness against adversarial attacks can be improved through the proposed approach in both centralized and decentralized environments. In \authone{the} future, we plan to extend our experiments to large-scale datasets such as ImageNet~\cite{russakovsky2015imagenet} and explore the application of AT beyond computer vision.}
\section*{Acknowledgment}
We acknowledge NSF for partially sponsoring the work under grants \#1620868 with its REU, \#2228562, and \#2236283. We thank Cyber Florida for a seed grant. 
\bibliographystyle{elsarticle-num} 
\bibliography{refs}
\end{document}